\pdfoutput=1

\documentclass[11pt]{article}

\usepackage[final]{acl}

\usepackage{times}
\usepackage{latexsym}

\usepackage[T1]{fontenc}

\usepackage[utf8]{inputenc}

\usepackage{microtype}

\usepackage{inconsolata}

\usepackage{graphicx}

\usepackage[utf8]{inputenc} 
\usepackage[T1]{fontenc}    
\usepackage{hyperref}       
\usepackage{url}            
\usepackage{booktabs}       
\usepackage{amsfonts}       
\usepackage{nicefrac}       
\usepackage{microtype}      
\usepackage{xcolor}         
\usepackage{graphicx}
\usepackage[misc]{ifsym}
\usepackage{multirow}
\usepackage{graphicx}
\usepackage{amsmath}
\usepackage{mathtools}
\usepackage{amsthm}
\usepackage{comment}
\usepackage{subfigure}
\usepackage{booktabs}
\usepackage{enumerate}
\usepackage{enumitem}
\usepackage{amssymb}
\usepackage{pifont}

\usepackage{wrapfig}
\usepackage{booktabs}
\usepackage{colortbl}

\usepackage{hyperref}
\usepackage{cleveref}

\newcommand{\blue}[1]{\textcolor{blue}{#1}}
\newcommand{\zhongwei}[1]{{\color{black}{#1}}}

\newcommand{\cheliu}[1]{{\color{black}{#1}}}

\title{MEIT: Multimodal Electrocardiogram Instruction Tuning on \\ Large Language Models for Report Generation}

\author{
Zhongwei Wan\textsuperscript{1},
Che Liu\textsuperscript{2},
Xin Wang\textsuperscript{1},
Chaofan Tao\textsuperscript{3},
Hui Shen\textsuperscript{1},
Jing Xiong\textsuperscript{3},\\
\textbf{Rossella Arcucci}\textsuperscript{2},
\textbf{Huaxiu Yao}\textsuperscript{4},
\textbf{Mi Zhang}\textsuperscript{1} \\
\textsuperscript{1}The Ohio State University \quad
\textsuperscript{2} Imperial College London\\ 
\textsuperscript{3}The University of Hong Kong \quad
\textsuperscript{4}University of North Carolina at Chapel Hill \\
\url{https://github.com/AIoT-MLSys-Lab/MEIT}
}

\begin{document}

\newcommand{\mz}[1]{\textcolor{red}{#1}}
\newcommand{\zw}[1]{{\color{black}{#1}}} 
\newcommand{\bruce}[1]{{\color{black}{#1}}} 
\maketitle

\begin{abstract}
Electrocardiogram (ECG) is the primary non-invasive diagnostic tool for monitoring cardiac conditions and is crucial in assisting clinicians. Recent studies have concentrated on classifying cardiac conditions using ECG data but have overlooked ECG report generation, which is time-consuming and requires clinical expertise. To automate ECG report generation and ensure its versatility, we propose the \textbf{M}ultimodal \textbf{E}CG \textbf{I}nstruction \textbf{T}uning (\textbf{MEIT}) framework, the \textit{first} attempt to tackle ECG report generation with LLMs and multimodal instructions. To facilitate future research, we establish a benchmark to evaluate MEIT with various LLMs backbones across two large-scale ECG datasets. Our approach uniquely aligns the representations of the ECG signal and the report, and we conduct extensive experiments to benchmark MEIT with nine open-source LLMs using more than 800,000 ECG reports. MEIT's results underscore the superior performance of instruction-tuned LLMs, showcasing their proficiency in \textit{quality report generation}, \textit{zero-shot capabilities}, \textit{resilience to signal perturbation}, and \zw{\textit{alignment with human expert evaluation}}.  These findings emphasize the efficacy of \textbf{MEIT} and its potential for real-world clinical application. Our code is released at \url{https://github.com/AIoT-MLSys-Lab/MEIT}.

\end{abstract}

\section{Introduction}
\label{sec:introduction}

Electrocardiogram (ECG) is the primary tool for heart disease diagnosis. In standard practice, cardiologists examine these ECG recordings and manually generate detailed diagnostic reports, which is a complex and time-consuming process. Recently, AI models have been developed to streamline ECG data analysis for the classification task~\citep{hu2023spatiotemporal,liu2023etp,DBLP:journals/corr/abs-2403-06659}, yet the automatic generation of reports from ECG recordings remains relatively underexplored.

Unlike image-based medical report generation tasks (e.g., radiology reports), ECG report generation faces unique challenges due to the concise, keyword-centric content of ECG reports, which contrasts with the more extensive anatomical descriptions in radiology. Directly transferring methods from radiology is hindered by the distinct nature of ECG signals and the limited semantic overlap with imaging data. Furthermore, there is still a lack of comprehensive benchmarks for evaluating the performance of ECG report generation.

%

Drawing on the versatility of LLMs and multimodal LLMs (MLLMs)~\citep{achiam2023gpt, touvron2023llama, wan2023efficient, wang2024iot, zhu2024dglf, wang2024svd, zhang2025, zhu2025} in various language tasks, in this work, we introduce MEIT, a \textbf{M}ultimodal \textbf{E}CG \textbf{I}nstruction \textbf{T}uning framework that extends the capability of large language models (LLMs) for the task of ECG report generation. 
%
By aligning ECG recordings with human instructions, MEIT produces clinically relevant reports and demonstrates zero-shot performance when transferring across different continents and data collection devices. 
Concretely, we construct a multimodal instruction dataset from publicly available ECG recordings and propose an efficient attention-based fusion method that incorporates ECG signals into the latent space of LLMs without adding new parameters.

We also introduce a comprehensive benchmark to evaluate ECG report generation across two large-scale datasets (with 20K and 800K ECG-report pairs), covering four tasks: (1) report generation quality, (2) zero-shot transfer across datasets, (3) robustness under signal perturbations, and (4) alignment with expert evaluation. We assess MEIT using ten open-source LLMs, demonstrating: \textbf{(1)} the superior performance of MEIT in ECG report generation and effective ECG representation learning; and \textbf{(2)} the strong transferability of instruction-tuned LLMs across diverse clinical domains.

In summary, our primary contribution is the MEIT framework, a novel method for automated ECG report generation and evaluation based on LLMs. It features a lightweight, attention-based fusion module for integrating ECG signals and text, along with a newly designed four-task benchmark for ECG report evaluation. Empirical results highlight MEIT’s ability to generate high-quality reports, perform robustly under data perturbations, and align closely with expert assessments, thereby paving the way for improved ECG interpretation and broader innovations in embedding biomedical signals into LLMs.

\section{Related Work}

\textbf{Medical Report Generation.} 
%
Our work is closely related to medical report generation, which has been extensively studied in the context of medical images. Existing methods can be grouped into three main categories: (1) template-based methods, such as HRGR~\citep{li2018hybrid} and CMAS~\citep{jing2017automatic}; (2) data integration and coherence methods, such as PPKED~\citep{liu2021exploring} and CA~\citep{ma2021contrastive}; and (3) cross-modal alignment methods, exemplified by~\citet{chen2022cross} and~\citet{qin2022reinforced}. 
However, these methods are designed for medical images and struggle with ECG signals due to their distinct temporal and waveform characteristics. 

\vspace{1mm}
\noindent \textbf{Instruction Tuning.}
Our work is also related to instruction tuning, which improves zero-shot learning in LLMs through task-specific instructions~\citep{zhang2023instruction, wang2023far}. Models like InstructGPT~\citep{ouyang2022training}, FLAN-PaLM~\citep{chung2022scaling}, and Alpaca~\citep{taori2023alpaca} leverage instruction data, including human feedback. Multimodal variants, such as LLaVA~\citep{liu2023visual}, MiniGPT-4~\citep{zhu2023minigpt}, and AnyMAL~\citep{moon2023anymal}, extend this to visual tasks.
However, these methods focus on natural images and are not directly applicable to ECG signals. 
To bridge this gap, we introduce a specialized instruction-tuning framework for ECG report generation.

\vspace{1mm}
\noindent \textbf{LLMs for ECG.}
Few studies have explored the use of LLM for the analysis of ECG signals~\citep{liu2023biosignal,qiu2023transfer,yu2023zero}. Some of them~\citep{liu2023biosignal,yu2023zero} convert ECG signals into text features before importing them into LLMs, bypassing crucial modality-specific characteristics. In addition, these works focus on the task of disease classification rather than report generation.
In contrast, our method directly processes ECG signals and focuses on the task of report generation.

\begin{figure*}[t]
\centering
\includegraphics[width=0.95\textwidth]{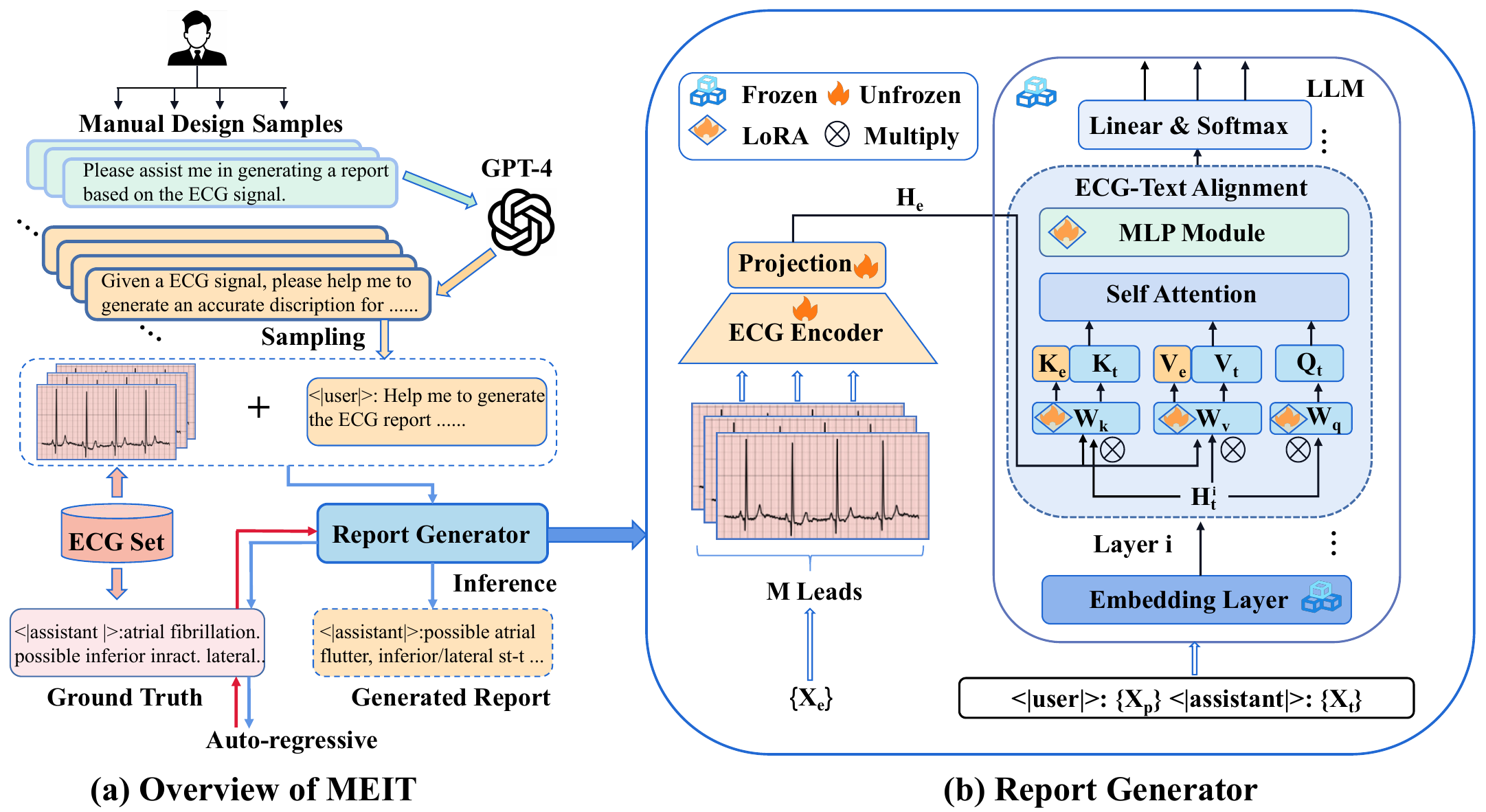}
\caption{\small{(a) Overview of MEIT; (b) Illustration of 
  the architecture of Report Generator. 
  }}
\label{fig:MEIT framework}
\end{figure*}

\section{MEIT}

\subsection{Preliminaries}

Electrocardiogram (ECG) measures the electrical activity of an individual's heart over time. An 12-lead ECG recording is a multivariate time series. It offers a multi-dimensional view, encompassing both spatial and temporal aspects of cardiac function. The 12 ECG leads include six limb leads (i.e., I, II, III, aVR, aVL, and aVF) that monitor arms and legs, providing frontal plane views, and six precordial leads (i.e., V1, V2, V3, V4, V5, and V6) that monitor chest, showing horizontal plane views. 

Formally, let $\mathbf{X}_e \in \mathcal{R}^{M \times T}$ denote an ECG recording, where $M$ represents the number of leads, and $T$ represents the signal length. Each ECG recording is associated with an ECG report $\mathbf{X}_{t}$ that describes and interprets the ECG recording. Thus, we denote each ECG recording-report pair as $\{\mathbf{X}_{e}, \mathbf{X}_{t}\}$. 

    


\subsection{Overview of MEIT}

Figure~\ref{fig:MEIT framework} (a) illustrates the proposed MEIT framework. \zw{In the data curation stage, we construct the ECG instruction tuning data, which includes instruction prompts, ECG recordings and the corresponding ground truth ECG reports. 
%
In the instruction tuning stage, the ECG instruction tuning data are fed into the Report Generator for training using an autoregressive approach. 
During inference, the instruction prompts and the ECG recordings are inputted into the Report Generator to generate the ECG reports. In the following sections, we describe each component in detail.}

\subsection{Data Curation}
\label{sec:instruction_data}
Given an ECG recording $\mathbf{X}_{e}$, our goal during inference is to generate an ECG report using an instruction prompt such as ``\textit{Given the ECG recording, please help me generate a report for this ECG recording:}". To achieve this goal, we aim to create a set of instruction tuning data to generate a response $\mathbf{\hat{X}}_{t}$ that semantically aligns with the ground truth $\mathbf{X}_{t}$. Since we cannot predict the exact instruction prompt that users will use, we need to ensure that our report generation process is robust enough to handle different prompts. To do so, we manually design a small set of prompt samples, and then utilize GPT-4~\citep{achiam2023gpt} to generate a larger prompt set by rephrasing, as shown in Figure~\ref{fig:MEIT framework} (a). 
For each ECG recording-report pair as $\{\mathbf{X}_{e}, \mathbf{X}_{t}\}$, we randomly select one instruction prompt $\mathbf{X}_{p}$ from the larger prompt set and create an instruction tuning template: \texttt{<|user|>}: \{$\mathbf{X}_{p}$, $\mathbf{X}_{e}$\} \texttt{<|assistant|>}: \{$\mathbf{X}_{t}$\} \texttt{</s>}, where \texttt{<|user|>} and \texttt{<|assistant|>} are added special tokens for tokenizer, and \texttt{</s>} is a stop sign for each response. 
This approach ensures that the generated response conveys the same meaning as the ground truth and remains adaptable to different instruction prompts. 
Following this strategy, we construct the ECG instruction tuning data using the PTB-XL~\citep{wagner2020ptb} dataset and the MIMIC-IV-ECG~\citep{mimicecg} dataset. 
\zhongwei{Some examples of ECG instruction tuning data are included in Appendix~\ref{appendix: sample visualization}.}

\subsection{Report Generation} 
\label{sec:ecg_generation_model}
Figure~\ref{fig:MEIT framework} (b) illustrates the architecture of the Report Generator. 
As shown, the Report Generator utilizes an ECG encoder to encode $\mathbf{X}_{e}$ into ECG embeddings and integrates them with the language embeddings extracted from the corresponding instruction tuning template through an ECG-text alignment module to autoregressively generate the ECG report. In this section, we describe the key components of the Report Generator in detail.

\vspace{1mm}
\noindent \textbf{ECG Encoder.} 
Since the ECG recording is high resolution in the time domain, it is vital to efficiently extract temporal features per lead before interaction with semantic embeddings inside the LLM backbone. 
Our default ECG encoder $\mathcal{F}_{e}(\cdot)$ consists of temporal convolution blocks to encode the ECG recordings into embeddings. 
Specifically, each temporal convolution block comprises several 1-D convolution layers, batch normalization layers, and ReLU activation layers, followed by average pooling. \zw{This design allows us to effectively capture temporal dependencies and reduce the complexity of the signal representations, ensuring that the model can quickly learn important temporal features efficiently. 
To further align the output dimension with the head dimension of the LLM backbone $\mathcal{F}_{l}(\cdot)$, we employ a non-linear projection layer $\mathcal{P}_{e}(\cdot)$ to generate the ECG embeddings:}

\begin{equation}
    \mathbf{H}_{e} = \mathcal{P}_{e}\left(\mathcal{F}_{e}\left(\mathbf{X}_{e}\right)\right),
\end{equation}
where $\mathbf{H}_{e} \in \mathcal{R}^{D_{h}}$, $D_{h}$ has the same dimension as the multi-head attention layers of LLMs. Note that our default ECG encoder is lightweight and is able to learn temporal patterns of ECG recordings without a long training period. \zhongwei{More details about the ECG encoder are included in Appendix~\ref{appendix: ECG Encoder}}. 

\vspace{1mm}
\noindent \textbf{ECG-Text Alignment.} 
Given the ECG and the text embeddings, we introduce an ECG-text alignment strategy to guide the LLM in aligning the ECG embeddings with the text embeddings. 
As shown in Figure~\ref{fig:MEIT framework} (b), given the ECG embeddings $\mathbf{H}_{e}$, the ECG-text alignment strategy incorporates $\mathbf{H}_{e}$ with the current hidden state $\mathbf{H}_{t}^{i}$ generated from the previous ($i-1$)th layer of the LLM backbone $\mathcal{F}_{l}\left(\cdot\right)$ for next-token prediction task. Here $\mathbf{H}_{t}^{i}$ is defined as:
\begin{equation}
   \mathbf{H}_{t}^{i} = \mathcal{F}_{l}^{i-1}\left(\left[\mathbf{X}_{p}, \mathbf{X}_{t}\right]\right),
\end{equation}
where $i$ is the current layer index. 

Traditional gated-attention fusion methods like  Flamingo~\citep{alayrac2022flamingo}, Memorizing Transformer~\citep{wu2022memorizing}, G-MAP~\citep{wan2022g}, and Q-former in BLIP-2~\citep{li2023blip} \bruce{require} additional trainable parameters and \zw{are designed for complex multi-stage alignment of rich semantic information (e.g., images). 
Different from them, our method provides a lightweight concatenation-based alignment strategy tailored to the ECG embeddings, enabling efficient learning of ECG semantic features via directly \bruce{injecting} the ECG embeddings with language context in the self-attention, while preventing potential catastrophic forgetting of general knowledge in LLMs.}

%
In our approach, each attention layer combines $\mathbf{H}_{e}$, generated from the ECG encoder and projector as a prefix condition, with $\mathbf{H}_{t}^{i}$, derived from the preceding layer. The fusion process is as follows:
\begin{equation}
   \operatorname{Self-Attn}\left(\mathbf{H}_{e}, \mathbf{H}_{t}^{i}\right) = \left[\operatorname{head}_{1},\ldots, \operatorname{head}_{k}\right]\mathbf{W}_{o},
\end{equation}
where $k$ represents the number of attention heads, and $\mathbf{W}_{o}$, a matrix in $\mathcal{R}^{kD_{h} \times D_{m}}$, serves as the projection matrix with $D_{m}$ denoting the hidden size of the LLM backbone. We replicate $\mathbf{H}_{e}$ for each head $k$ times, merging the ECG and language features in the sequence dimension. This is achieved through a shared projection of keys and values for each pattern. The fusion is then articulated as:
\begin{equation}
    \mathbf{K}_{m,j} = [\mathbf{K}_{e,j}, \mathbf{K}_{t,j}]^{\top},  \mathbf{V}_{m,j} = [\mathbf{V}_{e,j}, \mathbf{V}_{t,j}],
\end{equation}
\begin{equation}
    \operatorname{head}_j=\operatorname{Softmax}\left(\frac{\mathbf{Q}_{t,j} \mathbf{K}_{m,j}}{\sqrt{D_h}}\right) \mathbf{V}_{m,j},
\end{equation}
where $\mathbf{Q}_{t,j} = \mathbf{H}_{t,j}^{i} \mathbf{W}_{q,j}$,
$\mathbf{K}_{e,j}=\mathbf{H}_{e}\mathbf{W}_{k,j}$, and $\mathbf{K}_{t,j}=\mathbf{H}_{t,j}^{i}\mathbf{W}_{k,j}$, with a similar notation for $\mathbf{V}_{e,j}=\mathbf{H}_{e}\mathbf{W}_{v,j}$ and $\mathbf{V}_{t,j}= \mathbf{H}_{t}^{i}\mathbf{W}_{v,j}$. Concatenation is denoted by $[\cdot]$, and $\mathbf{K}_{m,j}$ and $\mathbf{V}_{m,j}$ symbolize the amalgamated features of query and key. $\mathbf{W}_{q,j}$, $\mathbf{W}_{k,j}$, and $\mathbf{W}_{v,j}$ in $\mathcal{R}^{D_{h} \times D_{h}}$ represent the projection matrices for query, key, and value for each head $j$, respectively. 
Such design allows for the efficient fusion of two modalities through causal attention, facilitating conditional generation without the need for additional parameter updates to align the ECG modality with the text modality. \zw{Ablation studies comparing with other fusion methods demonstrate the effectiveness and efficiency of our proposed lightweight alignment strategy.}
\zhongwei{More comparisons about ECG-text alignment and other fusion approaches are included in Table~\ref{tab: align ablaition}}. 

\subsection{Instruction Tuning}
During instruction tuning, we follow the instruction tuning template: \texttt{<|user|>}: \{$\mathbf{X}_{p}$, $\mathbf{X}_{e}$\} \texttt{<|assistant|>}: \{$\mathbf{X}_{t}$\} \texttt{</s>} and compute the autoregressive loss only on tokens after response tokens $\texttt{<assistant>}$, and use label loss masking to finetune the model, where we mask all tokens belonging to $\mathbf{X}_p$ and $\mathbf{X}_e$. 
To save computational resources and accelerate the convergence of instruction tuning, we use LoRA~\citep{hu2021lora} adapters for all linear layers of the LLM backbone $\mathcal{F}_{l}$ and freeze its backbone. Subsequently, given a sequence of ECG instruction data, we compute the probability of the target response $\mathbf{X_{t}}$ as an autoregressive function:
\begin{equation}
    p\left(\mathbf{X}_{t} \mid \mathbf{X}_{p}, \mathbf{X}_{e}\right) = \prod_{i=j}^L p_{\boldsymbol{\theta}}\left(\mathbf{x}_{t,i} \mid \mathbf{X}_{p}, \mathbf{X}_{e}, \mathbf{X}_{t,<i} \right),
\end{equation}
where $j$ is the start index after $\texttt{<assistant>}$, $\theta$ is the trainable parameters of LoRA and ECG encoder $\mathcal{F}_{e}$, and $\mathbf{X}_{t,<i}$ is the response tokens before the current generation $\mathbf{x}_{t,i}$.
\section{ECG Report Generation Benchmark}


\subsection{Datasets}

\noindent \textbf{MIMIC-IV-ECG.} 
MIMIC-IV-ECG~\citep{mimicecg} is the largest publicly available 12-lead ECG dataset. MIMIC-IV-ECG contains $800,035$ samples from $161,352$ subjects. Similar to PTB-XL, each sample includes a raw ECG recording and its report, sampled at $500$Hz for 10 seconds. The dataset is split into training, validation, and test sets at a ratio of 80\%:10\%:10\% respectively. 

\vspace{1mm}
\noindent \textbf{PTB-XL.}
We also use the PTB-XL dataset~\citep{wagner2020ptb}, which contains $21,837$ clinical 12-lead ECG recordings collected from $18,885$ patients. Each ECG recording is $10$ seconds long sampled at $500$Hz and has a corresponding report. 
We divide PTB-XL into training, validation, and test sets at a ratio of 70\%:10\%:20\%, respectively. 

%

\subsection{Models}

We use two smaller
language models and ten LLMs based on the
PEFT\footnote{https://github.com/huggingface/peft} library, which directly supports LoRA~\citep{hu2021lora} to construct the multimodal ECG report generation model described in Section~\ref{sec:ecg_generation_model}. 
The two smaller
language models are GPT2-Medium and GPT-Large~\citep{radford2019language}.
%
The ten LLMs are GPT-Neo~\citep{gpt-neo}, GPT-NeoX~\citep{black2022gpt}, GPT-J~\citep{gpt-j}, BLOOM~\citep{workshop2022bloom}, OPT~\citep{zhang2022opt}, LLaMA-1~\citep{DBLP:journals/corr/abs-2302-13971}, LLaMA-2-Instruct~\citep{touvron2023llama}, LLaMA-3-Instruct~\citep{touvron2023llama}, Mistral~\citep{jiang2023mistral}, and Mistral-Instruct.

\subsection{Evaluation Metrics}

We evaluate the performance using ten metrics: BLEU 1-4~\citep{papineni2002bleu}, METEOR~\citep{banerjee2005meteor}, ROUGE 1, 2 and L~\citep{lin2004rouge}, CIDEr-D~\citep{vedantam2015cider}, and BERTScore~\citep{zhang2019bertscore}. 
Specifically, BLEU 1-4 and METEOR assess machine translation quality, focusing on accuracy and fluency. ROUGE-L measures sentence fluency and structure, while ROUGE-1 and ROUGE-2 examine uni-gram and bi-gram overlaps. CIDEr-D evaluates the relevance and uniqueness of the generated ECG reports against a candidate set. Lastly, BERTScore assesses semantic similarity to the ground truth.

\begin{table*}[h]
    \centering
    \vspace{-7pt}
    \caption{\small{Natural language generation metric on MIMIC-IV-ECG. For model size, 'M' denotes the million level, and 'B' denotes the billion level. All checkpoints are downloaded from Hugging Face website. And all models have been fine-tuned using ECG instructions.
    The \colorbox{teal!15}{light teal} color indicates the second highest results, and
    \colorbox{teal!40}{heavy teal} color indicates the highest results.}}
    \label{tab: res mimic nlg}
    \scalebox{0.71}{
    \begin{tabular}{c|c|ccccccccc}
   \toprule[1.2pt]
\textsc{Models} & \textsc{Size} & \textbf{BLEU-1} & \textbf{BLEU-2} & \textbf{BLEU-3} & \textbf{BLEU-4} & \textbf{METEOR} & \textbf{ROUGE-L} & \textbf{ROUGE-1} & \textbf{ROUGE-2} & \textbf{CIDEr-D} \\
\midrule[1.2pt]
GPT2-Medium& 345M& 0.576 & 0.527 & 0.456 & 0.425 & 0.551 & 0.523 & 0.544 & 0.512 & 3.70 \\
GPT2-Large& 774M & 0.614 & 0.563 & 0.490 & 0.476 & 0.595 & 0.571 & 0.585 & 0.538 & 4.21 \\
\midrule
GPT-Neo &2.7B & 0.631 & 0.579 & 0.534 & 0.489 & 0.727 & 0.689 & 0.715 & 0.592 & 4.81 \\
GPT-NeoX &20B & 0.645 & 0.588 & 0.539 & 0.523 & 0.719 & 0.701 & 0.712 & 0.622 & 4.92 \\
GPT-J & 6B & 0.676 & 0.628 & 0.584 & 0.542 & 0.756 & 0.721 & 0.744 & 0.632 & 5.23 \\
BLOOM & 7B & 0.669 & 0.624 & 0.591 & 0.550 & 0.758 & 0.725 & 0.745 & 0.639 & 5.19 \\
OPT &6.7B & 0.673 & 0.616 & 0.598 & 0.532 & 0.755 & 0.732 & 0.743 & 0.631 & 5.32 \\
LLaMA-1 & 7B & 0.685 & 0.648 & 0.615 & 0.543 & 0.761 & 0.724 & 0.742 & 0.642 & 5.26 \\
Mistral &7B & 0.697 & 0.659 & 0.611 & 0.571 & 0.763 & 0.740 & 0.765 & 0.658 & 5.48 \\
\midrule
LLaMA-2-Instruct & 7B &0.706 & 0.662 & \cellcolor{teal!15}0.622 & \cellcolor{teal!15}0.581 & \cellcolor{teal!15}0.775 & 0.745 & \cellcolor{teal!15}0.768 & 0.664 & 5.55 \\
Mistral-Instruct& 7B & \cellcolor{teal!15}0.714 & \cellcolor{teal!15}0.665 & 0.619 & 0.576 & 0.768 & \cellcolor{teal!15}0.751 & 0.762 & \cellcolor{teal!15}0.667 & \cellcolor{teal!15}5.62 \\
LLaMA-3-Instruct & 8B & \cellcolor{teal!40}0.733 & \cellcolor{teal!40}0.686 & \cellcolor{teal!40}0.648 & \cellcolor{teal!40}0.610 & \cellcolor{teal!40}0.799 & \cellcolor{teal!40}0.773 & \cellcolor{teal!40}0.795 & \cellcolor{teal!40}0.686 & \cellcolor{teal!40}5.78 \\

\midrule[1.2pt]
\end{tabular}
}
\end{table*}

\begin{table*}[h]
    \centering
    \caption{\small{Natural language generation metric on PTB-XL. The \colorbox{teal!15}{light teal} color indicates the second highest results, and
    \colorbox{teal!40}{heavy teal} color indicates the highest results.}}
    
    \scalebox{0.71}{
    \begin{tabular}{c|c|ccccccccc}
   \toprule[1.2pt]
\textsc{Models}&\textsc{Size} & \textbf{BLEU-1} & \textbf{BLEU-2} & \textbf{BLEU-3} & \textbf{BLEU-4} & \textbf{METEOR} & \textbf{ROUGE-L} & \textbf{ROUGE-1} & \textbf{ROUGE-2} & \textbf{CIDEr-D} \\
\midrule[1.2pt]
GPT2-Medium &345M & 0.329 & 0.278 & 0.254 & 0.232 & 0.441 & 0.391 & 0.561 & 0.433 & 2.12 \\
GPT2-Large &774M & 0.437 & 0.395 & 0.355 & 0.320 & 0.575 & 0.481 & 0.652 & 0.527 & 3.25 \\
\midrule
GPT-Neo&2.7B & 0.474 & 0.449 & 0.398 & 0.373 & 0.602 & 0.486 & 0.674 & 0.595 & 3.70 \\
GPT-NeoX& 20B& 0.469 & 0.453 & 0.417 & 0.399 & 0.620 & 0.553 & 0.688 & 0.622 & 3.58 \\
GPT-J&6B & 0.485 & 0.452 & 0.428 & 0.405 & 0.656 & 0.550 & 0.662 & 0.613 & 3.72 \\
BLOOM & 7B & 0.491 & 0.462 & 0.427 & 0.415 & 0.665 & 0.580 & 0.678 & 0.605 & 3.80 \\
OPT &6.7B & 0.502 & 0.477 & 0.431 & 0.418 & 0.662 & 0.568 & 0.669 & 0.624 & 3.94 \\
LLaMA-1 &7B  & 0.514 &  \cellcolor{teal!15}0.485 & 0.465 & 0.430 & \cellcolor{teal!15}0.678 & 0.588 & 0.682 & 0.613 & 3.97 \\
Mistral& 7B & 0.486 & 0.475 & 0.446 & 0.421 & 0.673 & 0.591 & 0.697 & 0.634 & 3.98 \\
\midrule
LLaMA-2-Instruct& 7B  &  \cellcolor{teal!15}0.515 & 0.484 & \cellcolor{teal!15}0.469 & \cellcolor{teal!15}0.439 & 0.675 & \cellcolor{teal!15}0.594 & 0.698 & 0.624 & \cellcolor{teal!15}4.05 \\
Mistral-Instruct& 7B & 0.501 & 0.481 & 0.457 & 0.425 & 0.664 & 0.592 & \cellcolor{teal!15}0.700 &\cellcolor{teal!15}0.641 & 4.01 \\
LLaMA-3-Instruct & 8B  & \cellcolor{teal!40}0.539 & \cellcolor{teal!40}0.513 & \cellcolor{teal!40}0.494 & \cellcolor{teal!40}0.467 & \cellcolor{teal!40}0.698 & \cellcolor{teal!40}0.615 & \cellcolor{teal!40}0.725 & \cellcolor{teal!40}0.646 & \cellcolor{teal!40}4.45 \\

\bottomrule[1.2pt]
\end{tabular}
}
    \label{tab: res ptb nlg}
\end{table*}

\subsection{Tasks}

\noindent \textbf{Task \#1: Quality of the Generated Reports.} 
In our first task, we aim to assess the quality of the generated ECG reports using 10\% of PTB-XL and MIMIC-IV-ECG datasets as the test set. 
Specifically, this task examines how closely the generated reports match the ground truth reports in terms of structure and semantic meaning under various instructions and ECG recordings.
%

\vspace{1mm}
\noindent \textbf{Task \#2: Zero-shot Generalizability.} 
As the second task, to explore the generalizability of LLMs in domain transfer scenarios following ECG instruction tuning, we trained the models on 70\% of the instruction data from MIMIC-IV-ECG. 
\cheliu{Following this, we evaluated the models' zero-shot capabilities on the PTB-XL test set. It is important to note that the PTB-XL and MIMIC-IV-ECG datasets originate from different continents, Europe, and the United States, respectively, utilizing different devices and from different hospitals, across different time periods. Therefore, we consider these datasets to represent two separate domains. This distinction allows us to use the PTB-XL dataset to gauge our model's performance in zero-shot domain transfer effectively.}
We used the metrics BLEU-4, METEOR,  ROUGE-L, and CIDEr-D because of  limited space and calculated their average for model evaluation.

\vspace{1mm}
\noindent \textbf{Task \#3: Signal Perturbation Robustness.} In real-world clinical settings, ECG signals often contain some degree of noise.
To evaluate the robustness of MEIT against such noisy interference, we selected 10\% of the ECG samples from the MIMIC-IV-ECG test dataset. We then added Gaussian noise to these samples during the models' instruction-based inference process. 
For this evaluation, we used BLEU-4, METEOR, ROUGE-L, and CIDEr-D as metrics.

\vspace{1mm}
\noindent \textbf{Task \#4: Evaluation of Alignment with Human Expert Annotations.} 
Lastly, to evaluate the differences between the reports generated by MEIT and human expert annotations, we established specific evaluation criteria and utilized closed-source LLMs to conduct a professional assessment of both the generated reports and expert annotations.
\section{Experiments and Analysis}

\begin{table}[t]
    \centering
    \caption{\small{Semantic similarity between the generated ECG reports and ground truths is measured using BERTScore, denoted as P for Precision, R for Recall, and F-1 for the F-1 Score.}}
    \scalebox{0.68}{
    \begin{tabular}{c|ccc|ccc}
   \toprule[1.2pt]
    &  \multicolumn{3}{c}{\textsc{MIMIC-IV-ECG}} & \multicolumn{3}{c}{\textsc{PTB-XL}} \\
    \midrule[1.2pt]
\textsc{Models} & \textbf{P} & \textbf{R} & \textbf{F-1} & \textbf{P} & \textbf{R} & \textbf{F-1} \\
\midrule[1.2pt]
GPT2-Medium & 0.562 & 0.453 & 0.502 & 0.534 & 0.465 & 0.497 \\
GPT2-Large & 0.657 &  0.574 & 0.613 & 0.625 & 0.553 & 0.586 \\
\midrule
GPT-Neo & 0.723 &  0.633& 0.675 & 0.675 & 0.588 & 0.628 \\
GPT-NeoX & 0.719 & 0.638& 0.676 & 0.654 & 0.579 & 0.614 \\
GPT-J & 0.725 &  0.655& 0.688 & 0.689 & 0.622 & 0.654 \\
BLOOM & 0.734 &  0.684& 0.708 & 0.701 & 0.645 & 0.672 \\
OPT & 0.713 &  0.667& 0.689 & 0.712 & 0.648 & 0.678 \\
LLaMA-1 & 0.752 &  0.697& 0.723 & 0.725 & 0.657 & 0.689 \\
Mistral & 0.761 &  \cellcolor{teal!15}0.732& 0.746 & 0.711 & 0.664 & 0.687 \\
\midrule
LLaMA-2-Instruct & 0.764 & 0.725& 0.744 & 0.721 & \cellcolor{teal!15}0.668 & 0.693 \\
Mistral-Instruct& \cellcolor{teal!15}0.773 &  0.722& \cellcolor{teal!15}0.747 & \cellcolor{teal!15}0.730 & 0.661 &  \cellcolor{teal!15}0.694\\ 
LLaMA-3-Instruct&\cellcolor{teal!40} 0.798  &\cellcolor{teal!40} 0.745& \cellcolor{teal!40}0.771& \cellcolor{teal!40} 0.745&\cellcolor{teal!40} 0.682 & \cellcolor{teal!40}0.712 \\ 
\midrule[1.2pt]
\end{tabular}
}
    \label{tab: semantic}

\end{table}

\subsection{Experimental Setup}
%

%
We utilized PyTorch 2.1, transformers~\citep{wolf-etal-2020-transformers}, and accelerated on A100 GPUs with LLMs from Hugging Face~\citep{wolf2019huggingface} ranging from 2.7 to 70 billion parameters. 
For larger models, we used DeepSpeed\footnote{https://github.com/microsoft/DeepSpeed}. The training covered 5 epochs on MIMIC-IV-ECG and PTB-XL with a 2e-5 learning rate and 64 batch size, employing a linear optimizer with a 0.03 warm-up ratio. 
For text preprocessing, we initially remove all instances of the `nan' string and sentences that consist solely of numerical values. 
Subsequently, we discard any samples whose reports contain fewer than 5 tokens. 
For ECG encoder, we adopt random initialization.
Additionally, the default number of generated prompts from GPT-4 is 256. 

\subsection{Quality Evaluation}

\noindent \textbf{Performance on MIMIC-IV-ECG.} 
Table~\ref{tab: res mimic nlg} and \ref{tab: semantic} present the results of various types of language encoders $\mathcal{F}_{l}(\cdot)$ on MIMIC-IV-ECG. 
The results show that all ten LLMs perform better than the two smaller language models (GPT2-Medium and GPT2-Large) across all evaluation metrics. 
Notably, from GPT-Neo to Mistral-Instruct,  LLM-based backbones achieve a significant margin over SLMs in all metrics. For instance, compared to GPT2-Large, the METEOR score increases in the range of 0.132 to 0.18 from GPT-Neo to LLaMA-2, and Mistral-Instruct outperforms GPT2-Large with an improvement of 0.18 in the ROUGE-L score and 0.134 in the F-1 of BERTScore.
The observed performance underscores the adeptness of LLMs in generalizing from ECG data showcasing enhanced proficiency in aligning ECG data representations with corresponding textual information. This highlights the significant potential of LLMs in ECG-to-text generation. Particularly, LLaMA-2-Instruct, Mistral-Instruct, and LLaMA-3-Instruct surpass their counterparts in most evaluative metrics, suggesting that models pre-tuned with general instructions are more adept at learning ECG-text alignment.


\vspace{1mm}
\noindent \textbf{Performance on PTB-XL.}  As shown in Table~\ref{tab: res ptb nlg}, the models exhibit reduced performance on PTB-XL compared to MIMIC-IV-ECG, which is attributable to the smaller scale of the instruction data in PTB-XL. 
This underscores the importance of data scale in enhancing instruction-based ECG report generation. Moreover, similar to the MIMIC-IV-ECG results, all LLM-based models show significant improvement over SLMs. Specifically, LLaMA-2 surpasses GPT2-Large by 0.134 in the BLEU-3 metric, while LLaMA-1 achieves a 0.103 improvement in the METEOR score. The overall experimental results also reveal that Mistral-Instruct, LLaMA-2-Instruct, and LLaMA-3-Instruct are consistently the top two performers across most metrics because of their strong general instruction-following capabilities.


\begin{table*}[t]
    \centering
    \caption{\small{Evaluation results of LLaMA-2-Instruct and LLaMA-3-Instruct against human expert-annotated ground-truth reports. Each dimension is scored on a scale of 1 to 5.}}
    \scalebox{0.82}{
    \begin{tabular}{c|c|c|c|c}
       \midrule[1.2pt]
        \textbf{Model} & \textbf{Medical Terminology Accuracy} & \textbf{Logical Consistency} & \textbf{Completeness} & \textbf{Diagnostic Accuracy} \\
        \midrule
        LLaMA-2-Instruct & 4.25 & 4.11 & 3.72 & 3.60 \\
        \midrule
        LLaMA-3-Instruct &  \textbf{4.52} & \textbf{4.38} & \textbf{4.01} & \textbf{3.98} \\
        \midrule[1.2pt]
    \end{tabular}}
    \label{tab:evaluation-results}
 
\end{table*}

\begin{figure}[t]
\centering
\includegraphics[width=0.5\textwidth]{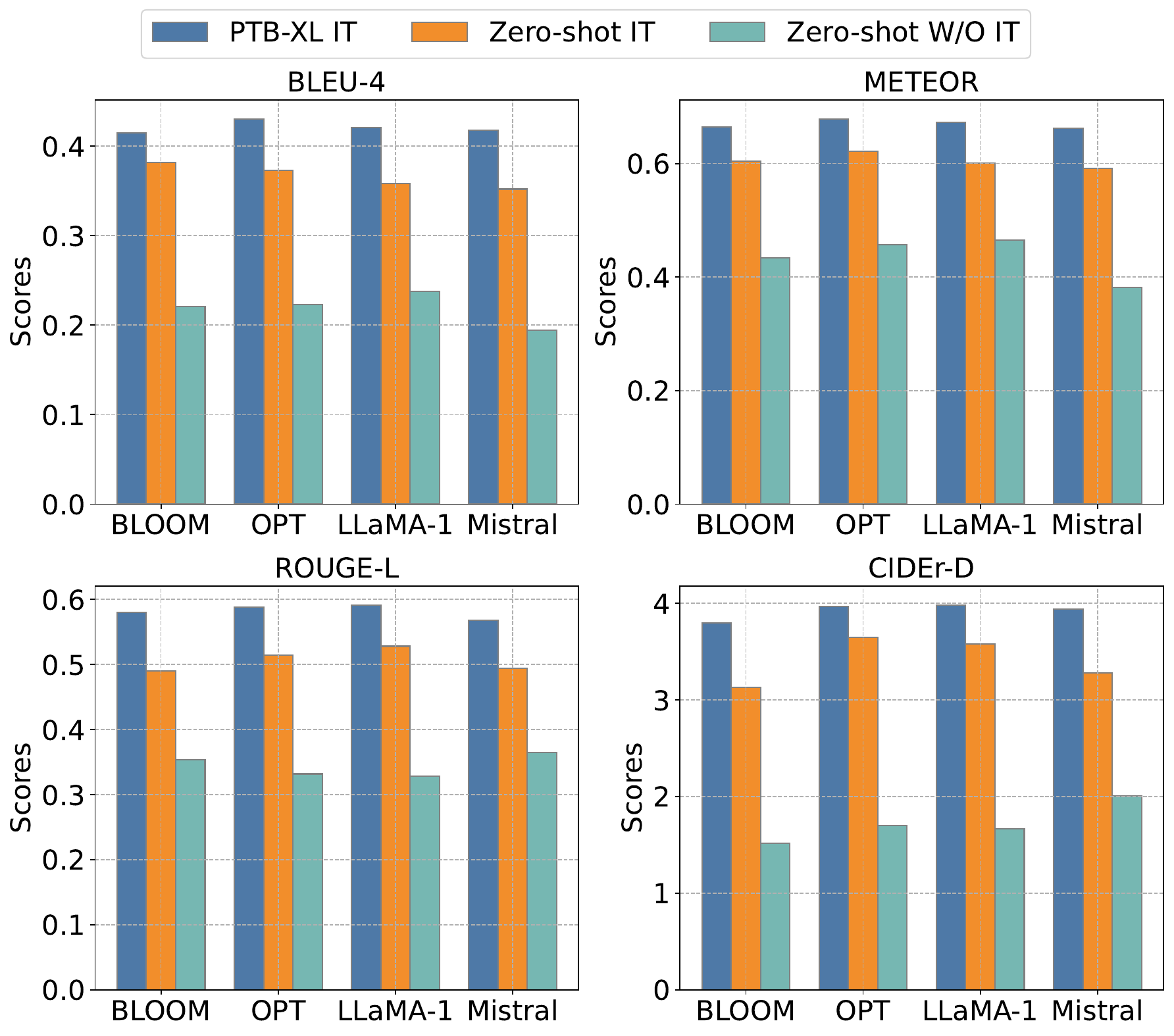}
\caption{\small{Zero-shot performance on PTB-XL dataset. ``IT'' denotes instruction tuning.}}
\label{fig:zeroshot}
\end{figure}
\subsection{Zero-shot Evaluation in Domain Transfer}

Although both PTB-XL and MIMIC-IV-ECG datasets are time-series data, they differ significantly in several aspects, including population (European vs. American), diverse collection devices, continents (Europe vs. US), protocols, and hospitals. These differences introduce substantial medical domain gaps~\citep{bilheimer2010data, ross2020influence}. 
In Figure \ref{fig:zeroshot}, we present the evaluation of the zero-shot learning capabilities of various LLMs, which is trained on the MIMIC-IV-ECG dataset and then tested on PTB-XL (unseen dataset). 
The assessed models include BLOOM, OPT, LLaMA-1, and Mistral. 
Firstly, all selected LLMs undergo instruction tuning on the MIMIC-IV-ECG train set, followed by zero-shot testing on the PTB-XL test set verified by human experts, denoted as \textsc{Zero-shot IT}. 
We also measure the performance of each model in report generation without prior ECG-specific instruction tuning, denoted as \textsc{Zero-shot W/O IT}. 
\textsc{PTB-XL IT} represents training on the PTB-XL train set and then evaluated on the PTB-XL test set.
Notably, although \textsc{Zero-shot IT} shows a slight degradation compared to \textsc{PTB-XL IT}, the results still indicate a variance in the model's ability to generalize to an unseen dataset with instruction tuning (IT), compared to \textsc{Zero-shot W/O IT}. 
The involvement of ECG instruction tuning on MIMIC-IV-ECG enables the models to achieve superior zero-shot performance on the unseen PTB-XL dataset, indicating the necessity of instruction tuning in enhancing the models' zero-shot ability on unseen datasets in ECG report generation.


\subsection{Robust Analysis with Perturbed ECG Recordings}

\begin{figure}[t]
\centering
\begin{minipage}[t]{0.85\linewidth}
    \centering
    \includegraphics[width=\linewidth]{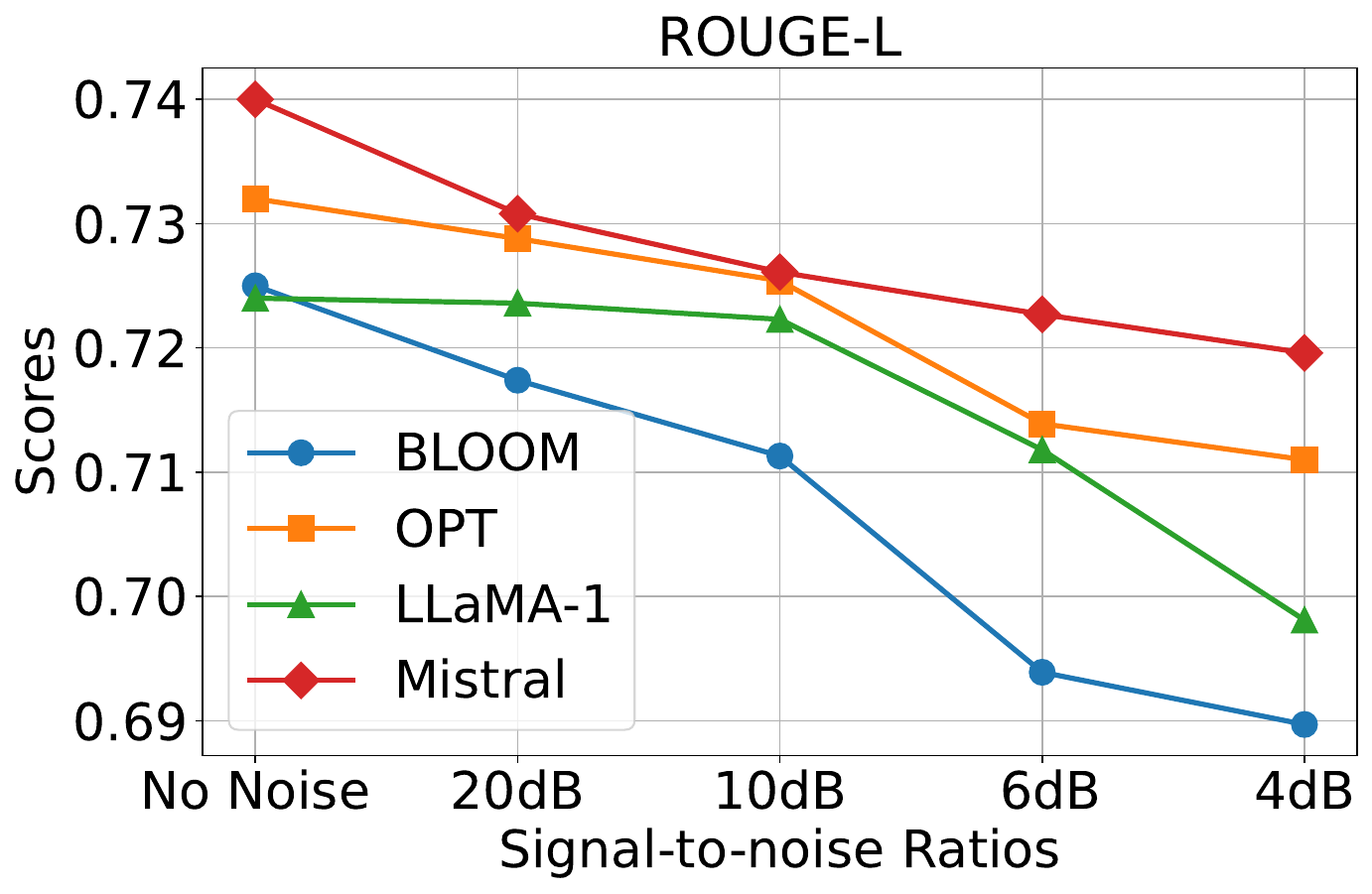}
    \vspace{1mm}
\end{minipage}

\begin{minipage}[t]{0.85\linewidth}
    \centering
    \includegraphics[width=\linewidth]{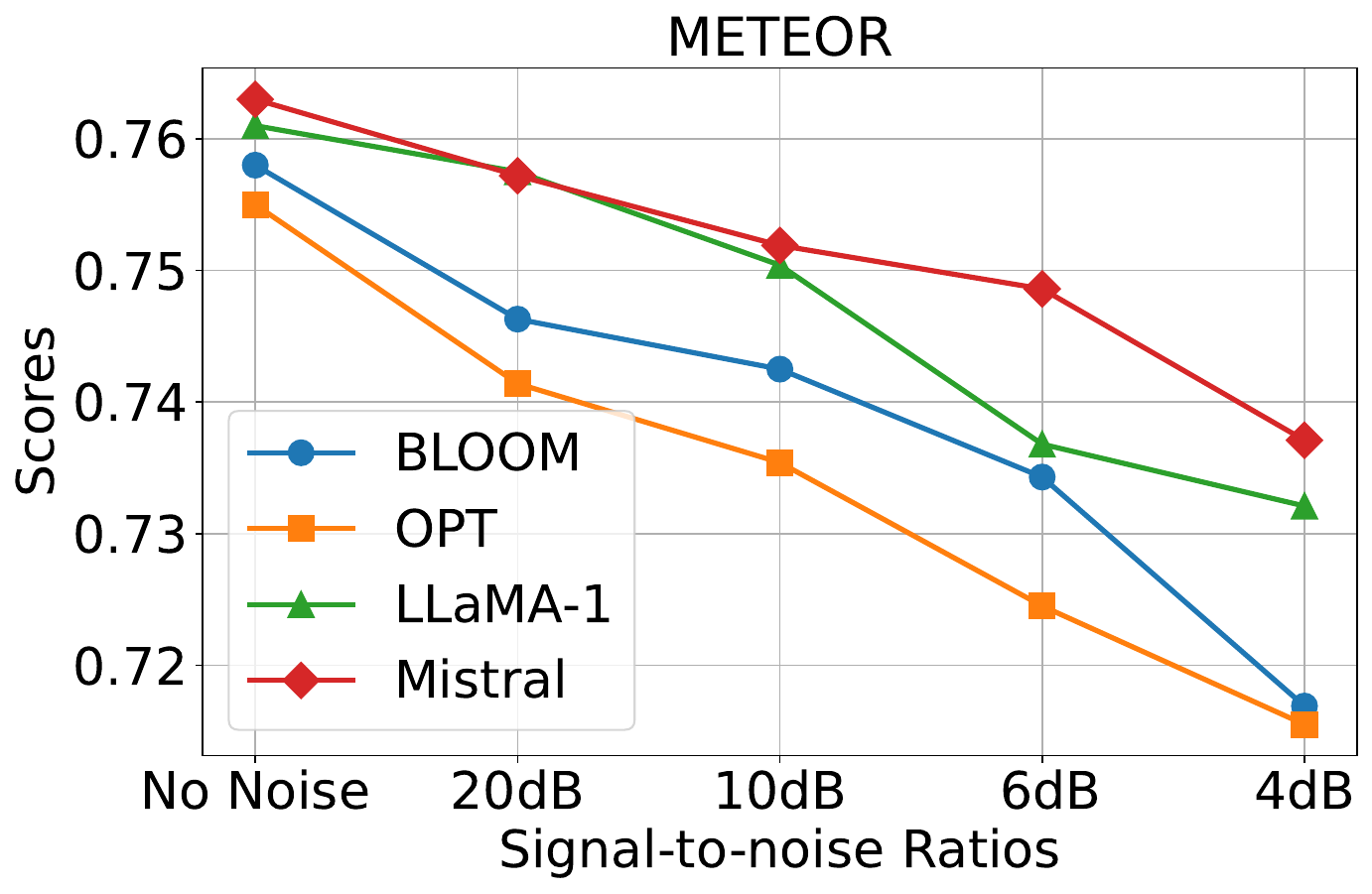}
\end{minipage}

\caption{\small{Signal perturbation robustness analysis on various LLMs.}}
\label{fig:robutness}
\end{figure}

In a noise stress evaluation \citep{wang2019adversarial}, we added Gaussian noise to ECG signals at signal-to-noise ratios (SNRs) of 0.05, 0.1, 0.15, and 0.2 during testing to assess model robustness. Our experiments utilized four LLM architectures: BLOOM, OPT, LLaMA-1, and Mistral, each trained on clean ECG signals from the MIMIC-IV-ECG training set and tested on corresponding noise-added signals from its test set. The results, illustrated in Figure \ref{fig:robutness}, show a performance decline in all LLMs as SNR decreased, highlighting the significant interference of ECG noise. Furthermore, as shown in Table \ref{tab: res mimic nlg}, Mistral also excelled in tests on noise-free datasets, suggesting a synergistic effect between clean and noisy test sets. The results demonstrate Mistral's strong resistance to perturbations. Even with more severe noise, it maintained robustness regarding ROUGE-L and METEOR metrics. 

\subsection{Evaluation of Alignment with Human Expert Annotations}

\zw{In Table~\ref{tab:evaluation-results}, we conducted an evaluation of model-generated ECG reports from ECG instruction-tuned versions of LLaMA-2 and LLaMA-3 against 500 ground-truth reports, meticulously annotated by human medical experts. These test annotated data were randomly sampled from the PTB-XL dataset, with all selected reports carefully reviewed and validated by human experts. Each model-generated report was compared with these expert-annotated reports using gpt-4o\footnote{https://platform.openai.com/docs/models/gpt-4o}, which assessed quality across four dimensions: \textbf{Medical Terminology Accuracy}, \textbf{Logical Consistency}, \textbf{Completeness}, and \textbf{Diagnostic Accuracy}, on a scale of 1 to 5. To evaluate these reports, we employed the following prompt template, which guided GPT-4o's scoring process across the defined dimensions, as shown in Table~\ref{tab:prompt-template} in Appendix. This prompt template ensures that GPT-4o evaluates the reports in a structured and consistent manner, highlighting both strengths and weaknesses of the model-generated reports in comparison to human expert annotations. }\zw{The results indicate that the LLaMA-3 model, with an average Diagnostic Accuracy score of 3.85, closely matches the quality of the human expert annotations, whereas the LLaMA-2 model scored 3.60. This evaluation underscores the effectiveness of using human expert annotations from the PTB-XL~\citep{wagner2020ptb} dataset as a rigorous benchmark for assessing the models' ability to generate clinically reliable ECG reports.}


\section{Conclusion}
\label{sec:conclusion}

%

In this paper, we introduced MEIT, a framework for generating instruction-following data to train a multimodal LLM that can produce ECG reports based on human instructions. We also proposed an effective method for aligning ECG and report representations across various open-source LLMs, demonstrating strong performance on both the MIMIC-IV-ECG and PTB-XL datasets across multiple tasks. Additionally, we established a comprehensive benchmark for ECG instruction-following in report generation, providing a standardized evaluation for future research. Although this work primarily focuses on ECG signals, it serves as a foundational step in applying instruction-tuning to biomedical signals. For future research, we aim to extend our framework and benchmark to other medical domains, such as EEG, with the hope of driving further progress in developing more capable medical-signal LLMs.

\section{Limitations}

Our MEIT framework first attempts to address automatic ECG report generation through ECG instruction tuning, establishing the first comprehensive benchmark for this process with mainstream LLMs. In this work, we mainly focus on generating reports from multimodal ECG instructions using LLMs. However, the generated results are not fully explainable or controllable, even though the generation procedure is transparent and trackable. This is because the underlying theory of LLMs remains largely unexplored, necessitating further investigation to ensure the quality and safety of the generated content. In the future, we aim to enable LLMs to utilize external, expert-verified knowledge databases, such as clinical protocols and medical textbooks, to enhance the explainability of the generated ECG reports.

\section{Acknowledgment}

We thank the reviewers and ACs for their helpful comments.

\bibliography{main}

\appendix
\appendix
\twocolumn
\section{Appendix.}

\subsection{Further Analysis of MEIT}
\label{sec: analysis}

\begin{figure}[h]
    \centering
    \includegraphics[width=0.45\linewidth]{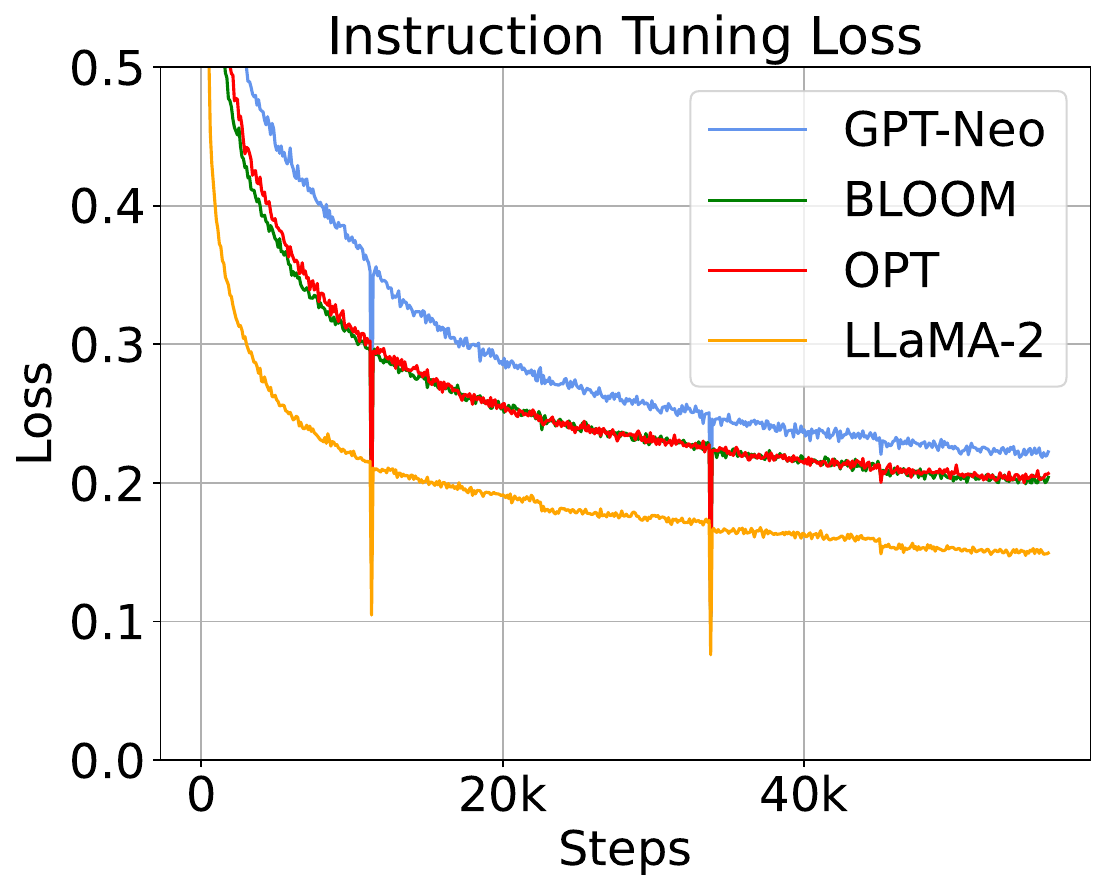} 
    \hspace{0.01\textwidth} 
    \includegraphics[width=0.45\linewidth]{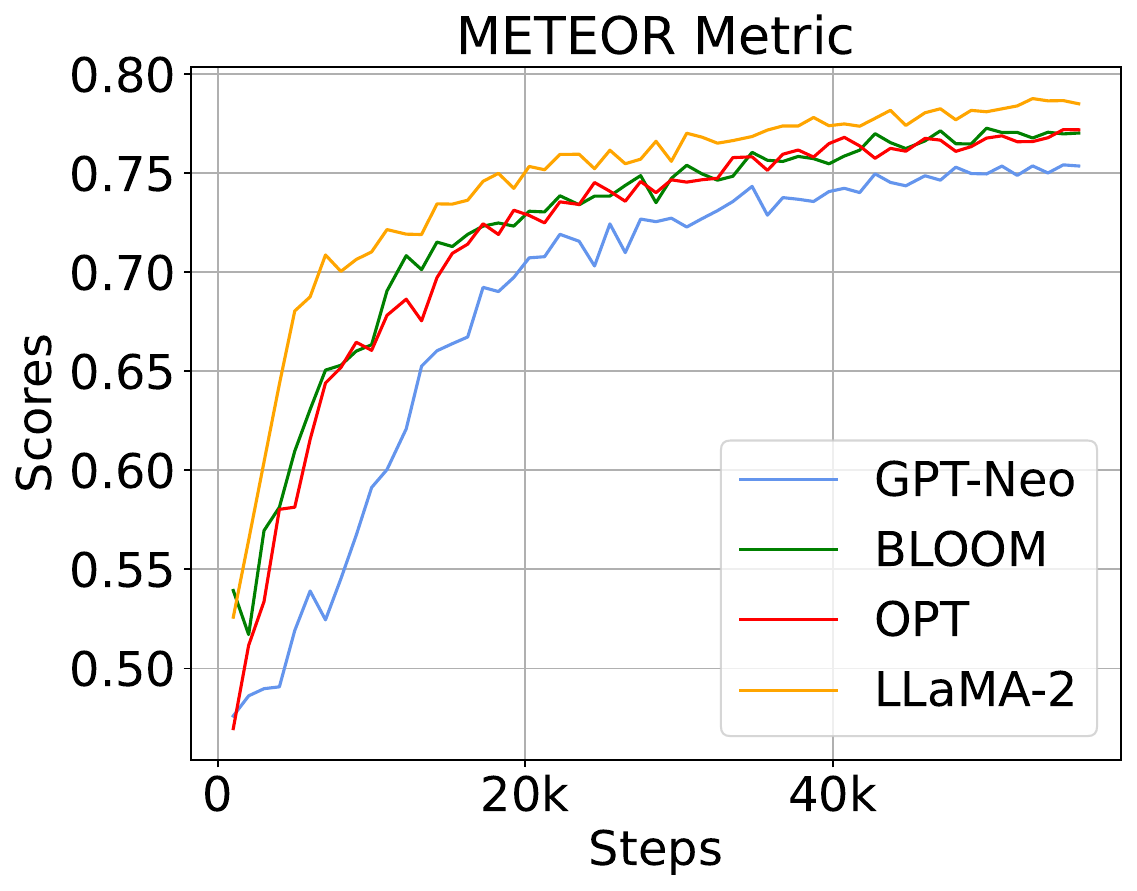} 
    \caption{\small{Visualizations of instruction tuning loss and METEOR score.}}
    \label{fig:loss_meteor}
\end{figure}

\noindent \textbf{Instruction Tuning Visualization.}
Figure~\ref{fig:loss_meteor} compares the convergence curves of the instruction tuning loss and the METEOR score between GPT-Neo (2.7B), BLOOM (7B), OPT (6.7B), and LLaMA-2 (7B) on the MIMIC-IV-ECG train and validation datasets. We observe that larger models with more parameters can converge to a more minor loss and achieve higher performance on the METEOR score. Notably, an increase in model size correlates with higher performance and lower loss, suggesting that larger models have the potential for better performance.

\begin{table*}[t]
    \centering
    \caption{\small{Performance comparison of the proposed concatenated-fusion method and other mainstream fusion variants. We evaluate these methods on the MIMIC-IV-ECG dataset, using BLEU-4, METEOR, ROUGE-L, and CIDEr-D metrics. We take LLaMA-1 7B as the LLM backbone here. 
    \colorbox{teal!40}{heavy teal} color indicates the highest results.}}
    \scalebox{0.75}{
    \begin{tabular}{c|c|cccc}
   \toprule[1.2pt]
\textsc{Framework}&\textsc{Method} & \textbf{BLEU-4} & \textbf{METEOR} & \textbf{ROUGE-L}  & \textbf{CIDEr-D} \\
\midrule[1.2pt]
 LLaVA& Straightforward input & 0.529 & 0.737 & 0.712& 4.99\\
 Flamingo& Trainable cross-attention &0.527 & \cellcolor{teal!40}0.768& 0.715& 5.11\\
 \midrule
 MEIT& Concatenated-fusion & \cellcolor{teal!40}0.543 & 0.761 & \cellcolor{teal!40}0.724 & \cellcolor{teal!40}5.26\\
\bottomrule[1.2pt]
\end{tabular}

}
    
    \label{tab: align ablaition}
\end{table*}

\noindent \textbf{Analysis of ECG Modality Alignment.}
To study the effectiveness of our proposed concatenated-fusion method for ECG modality alignment, we compare it with other fusion approaches such as direct input in LLaVA~\citep{liu2023visual} and additional trainable cross-attention layer in Flamingo~\citep{alayrac2022flamingo}. For straightforward input, we follow the design of LLaVA by directly concatenating the ECG encoder's output embeddings with the sentence's embeddings before inputting them into the LLM backbones. For the second comparison method, we follow Flamingo by adding a trainable cross-attention layer within the attention block. From Table~\ref{tab: align ablaition}, we observe that the Concatenated-fusion method outperforms the trainable cross-attention method of Flamingo in most metrics and is consistently superior to the Straightforward input method of LLaVA. Consequently, the concatenated fusion is more effective for the LLM backbone's alignment with fine-grained ECG patterns without necessitating additional trainable parameters.

\begin{figure}[h]
    \centering
    \includegraphics[width=\linewidth]{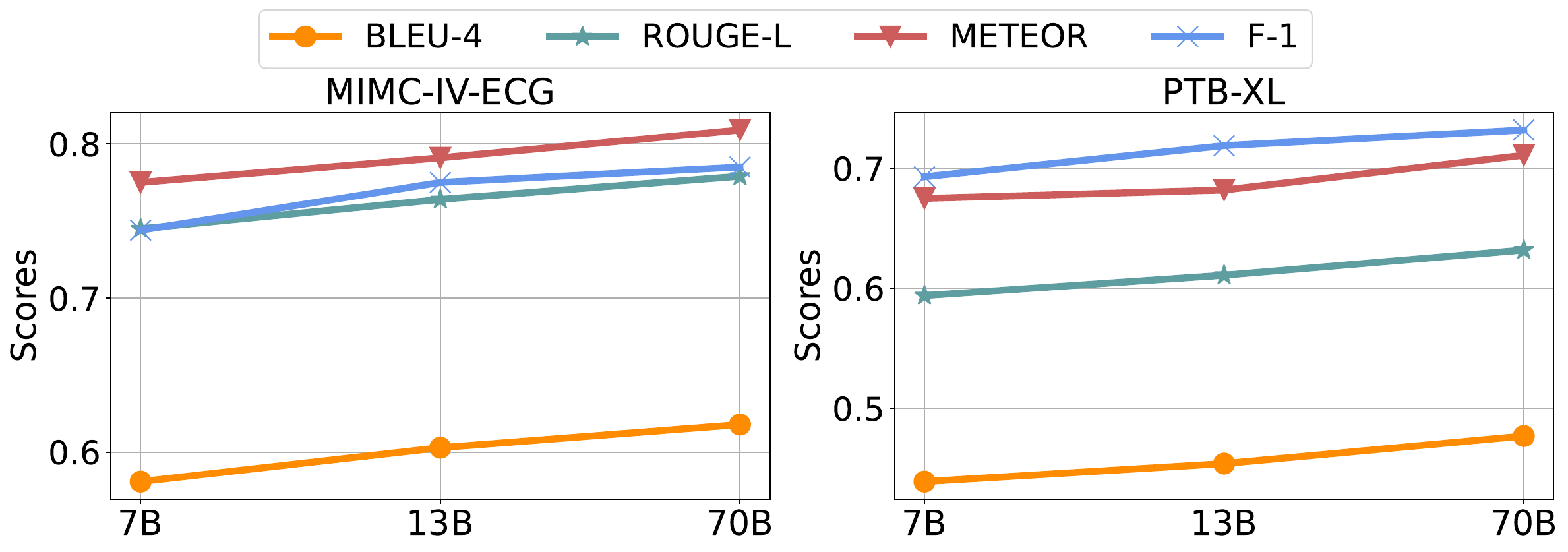}
    \caption{\small{Model scaling performance on MIMIC-IV-ECG and PTB-XL.}}
    \label{fig:scaling_test}
\end{figure}

\noindent \textbf{Scalability Analysis.}
To investigate whether ECG instruction tuning on larger-scale models yields better results, we validated LLaMA-2 models of \zhongwei{7B, 13B, and 70B} parameter sizes on both MIMIC-IV-ECG and PTB-XL datasets. 
As depicted in Figure~\ref{fig:scaling_test}, an upward trend in all evaluation metrics is observed with a gradual increase in model size.

However, it is noteworthy that the gains in performance associated with increasing model size are not particularly significant. For example, the F-1 score for the 70B model on the PTB-XL dataset exhibits a marginal increase of 0.02 over the 13B model. Similarly, on the MIMIC-IV-ECG dataset, the 70B model's F-1 score is only 0.01 higher than that of the 13B model. Therefore, we conjecture that enhancing both data scale and model size concurrently is necessary to achieve superior performance~\citep{wei2022emergent}.

 \begin{figure*}[!htbp]
\centering
\includegraphics[width=1.0\textwidth]{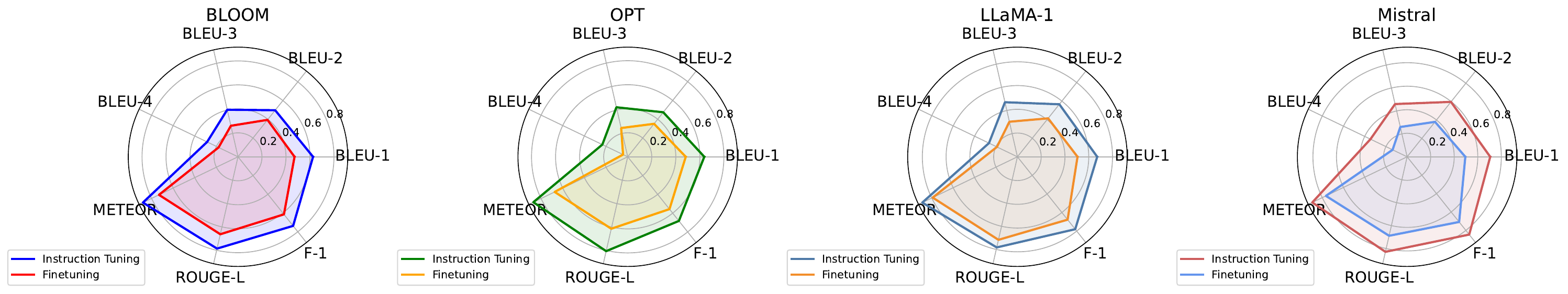}
\caption{\small{Ablation Study of ECG Instruction Tuning on MIMIC-IV-ECG Dataset.}}
\label{fig:IT_ablation}

\end{figure*}

\begin{figure}[!htbp]
\centering
\includegraphics[width=0.46\textwidth]{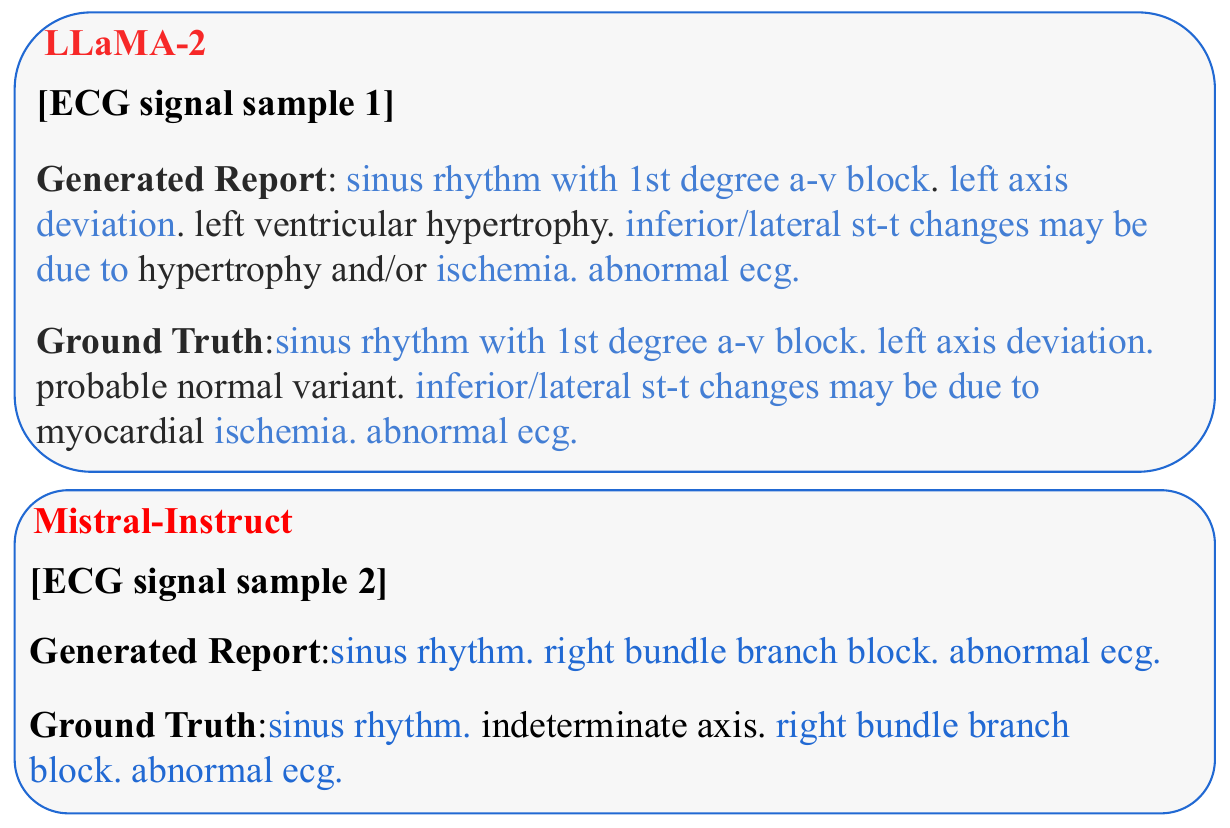}
\caption{\small{Examples of ECG reports generated by LLaMA-2 and Mistral-Instruct. We highlight the consistent information between the generated reports and the ground truths with \blue{blue} color. }}
\label{fig:samples}
\vspace{-1mm}
\end{figure}

\noindent \textbf{Ablation Study on ECG Instruction Tuning.}
We conducted an ablation study to evaluate instruction tuning's impact on aligning ECG signals with report representations. Utilizing LLMs such as BLOOM, OPT, LLaMA-1, and Mistral without instruction tuning, 
we allowed direct learning from ECG signals. The findings, illustrated in Figure~\ref{fig:IT_ablation}, indicate a significant performance drop across all metrics without instruction tuning, particularly in Mistral. This underscores instruction tuning's superiority in enhancing LLMs' generalization to new tasks/data over direct fine-tuning~\citep{ouyang2022training}.



\begin{table*}[h]
    \centering
    \caption{\small{Prompt template used for GPT-4o evaluation. This prompt guided the model's evaluation of generated ECG reports.}}
    \scalebox{0.75}{
    \begin{tabular}{|p{15cm}|}
        \midrule[1.2pt]
        \cellcolor{gray!15}
        \textbf{Prompt Template for GPT-4o Evaluation} \\
        \midrule
        You are an expert in Electrocardiogram (ECG) text evaluation. Your task is to assess the quality of a generated ECG report by comparing it to a real, expert-annotated ECG report. \\
        \textbf{Generated ECG Report}: \{Generated\_Report\} \\
        \textbf{Real ECG Report}: \{Real\_Report\} \\
        Please evaluate the generated report based on the following criteria: \\
        \midrule
        1. \textbf{Medical Terminology Accuracy}: Does the generated report use correct and appropriate ECG signal terms? \\
        2. \textbf{Logical Consistency}: Is the information presented in a logical and medically sound order? \\
        3. \textbf{Completeness}: Does the report include all necessary details that would be present in a real ECG report, such as heart rhythm, rate, and any abnormalities? \\
        4. \textbf{Diagnostic Accuracy}: Are the diagnoses and interpretations in the generated report accurate and consistent with the expert-annotated report? \\
        Please provide a detailed analysis and score each criterion on a scale of 1 to 5 (1 = Poor, 5 = Expert-Level). \\
        \midrule[1.2pt]
    \end{tabular}
    }
    \label{tab:prompt-template}
    \vspace{-10pt}
\end{table*}

\noindent \textbf{Qualitative Results.}
In Figure~\ref{fig:samples}, we randomly select two samples generated 
by MEIT using LLaMA-2 and Mistral-Instruct as the LLM backbones. 
The consistent key information, highlighted in blue, indicates that both models have successfully learned important patterns from the ECG signals. Overall, the models' results align with the ground truth, accurately identifying cardiac abnormalities from the ECG signals. Furthermore, both models provide detailed explanations of abnormal ECG signal details, such as `\blue{ischemia}' from sample 1 and `\blue{right bundle branch block}' from sample 2. These generated reports demonstrate the efficacy of our method.

\subsection{Hyper-parameters of ECG Instruction Tuning}
\label{appendix: Hyper-parameters}

\begin{table*}

\begin{minipage}[t]{0.5\textwidth}
\makeatletter\def\@captype{table}
\centering
\caption{Hyper-parameters of ECG instruction tuning for all LLM backbones. }
\scalebox{0.99}{
\begin{tabular}{lc}
        \midrule[1.2pt]
        \textbf{Hyperparameters} &  \\
        \midrule
Mixed precision & bf16 \\
Instruction tuning epochs & 5 \\
LoRA alpha & 64 \\
LoRA rank & 128 \\
LoRA dropout & 0.1 \\
Total batch size & 64 \\
Gradient accumulation & 2 \\
Maximum sequence length & 256 \\
Learning rate & 2e-5, 1e-4 \\
Learning rate Optimizer & AdamW \\
Schedule & linear \\ 
Warm-up ratio & 0.03 \\
Weight decay & 0.0 \\
        \midrule[1.2pt]
        \end{tabular}
}
\label{tab: hyperparameter}
\label{tab: hyperparameter for MLM}
\end{minipage}
\begin{minipage}[t]{0.5\textwidth}
\makeatletter\def\@captype{table}
\centering
\caption{ECG dimension of different language models.}
\scalebox{0.90}{
\begin{tabular}{lc}
        \midrule[1.2pt]
        \textsc{Models} &  \textbf{ECG Dimension}\\
        \midrule
        GPT2-Medium &  64 \\
        GPT2-Large     & 64   \\
        GPT-Neo  & 128   \\
        GPT-NeoX & 96 \\
        GPT-J & 256   \\
        BLOOM & 128 \\
        OPT & 128  \\
        LLaMA-1 &  128  \\
        Mistral & 128 \\
        LLaMA-2 & 128 \\
        Mistral-Instruct & 128 \\ 
        \midrule[1.2pt]
        \end{tabular}
}
\label{tab: ECG dim}
\end{minipage}
\end{table*}

In this study, we implement the Low-Rank Adaptation (LoRA)~\citep{hu2021lora} technique for efficient fine-tuning, specifically applied to ECG instruction tuning. As detailed in Table~\ref{tab: hyperparameter} provided, we utilize mixed precision at bf16 for enhanced computational efficiency. Our models undergo instruction tuning over 5 epochs, with LoRA parameters set at an alpha of 64 and a rank of 128, accompanied by a dropout rate of 0.1. The total batch size is 64, with a gradient accumulation factor of 2. The maximum sequence length is constrained to 256 tokens.
Additionally, we adopt a learning rate with 2e-5 for GPT-NeoX and 1e-4 for the other models, optimized using the AdamW algorithm. The learning rate follows a linear schedule with a warm-up ratio of 0.03. We set the weight decay to 0.0. 

Moreover, as shown in Table~\ref{tab: ECG dim}, we detail the ECG embedding dimensions for various language models, highlighting their approach to ECG data encoding. GPT2-Medium and GPT2-Large feature ECG dimensions 64, while GPT-Neo, BLOOM, OPT, LLaMA-1, Mistral, LLaMA-2, and Mistral-Instruct use a dimension of 128. GPT-NeoX employs a dimension of 96, and GPT-J notably uses the largest dimension of 256. These dimensions, reflecting each model's head dimension design, illustrate diverse strategies in ECG data processing across different models.

\subsection{More Details of ECG Encoder}
\label{appendix: ECG Encoder}

\noindent \textbf{Projection Layer}  For the design of the projection layer within the ECG encoder, we adopt a non-linear approach similar to CLIP~\citep{radford2021learning} and Med-UniC~\citep{wan2024med}. Specifically, in our experiments, we employ two consecutive linear layers, each followed by BatchNorm1d\footnote{https://pytorch.org/docs/torch.nn.BatchNorm1d.html}. Besides, ReLU serves as the activation function between the two linear layers. The default settings for input and hidden layers dimensions are set to 2048 in our experiment.

\noindent \textbf{Parameter Size Analysis}  To demonstrate the ECG encoder's lightweight design, we analyzed its trainable parameters during instruction tuning and total parameters during inference, using the LLaMA-1 7B model for illustration (Table~\ref{tab: params compare}). The analysis reveals the ECG encoder's trainable parameters are substantially fewer than those of the LoRA adapter in the LLM backbone during instruction tuning, and its parameter share of the overall framework is minimal for inference, underscoring its efficiency.

\begin{table*}[h]
    \centering
    \caption{Comparisons of results with and without supervised manner. We take LLaMA-2-Instruct as the LLM backbone here. 
    \colorbox{teal!40}{heavy teal} color indicates the highest results.}
    \scalebox{0.75}{
    \begin{tabular}{c|c|cccc|cccc}
   \toprule[1.2pt]
   \textsc{Methods} & \textsc{Size} & \multicolumn{4}{c}{\textsc{MIMC-IV-ECG}} & \multicolumn{4}{c}{\textsc{PTB-XL}}\\
   \midrule[1.2pt]
& & \textbf{BLEU-4} & \textbf{METEOR} & \textbf{ROUGE-L}  & \textbf{CIDEr-D} & \textbf{MTA} & \textbf{MTA} & \textbf{LC} & \textbf{DA}\\
\midrule[1.2pt]
   Vision Mamba&  86M	 & 	0.548 &  0.737& 0.715&  5.58& 3.78	& 3.88	& 3.61& 3.50\\
   Vision Transformer& 	98M &  0.592&  0.815& 0.772& 5.67  & 4.33 & 4.15&  \cellcolor{teal!40}4.12& 3.78 \\
   Vision Transformer (SSL)& 98M & 0.581 &  \cellcolor{teal!40}0.822 & 0.766&  5.75& 4.42 & 4.28& 3.85&  3.85\\
   1-D Temporal Conv (Ours)& 20.4M &   \cellcolor{teal!40}0.610&  0.799 &   \cellcolor{teal!40}0.773&  \cellcolor{teal!40}5.78 &  \cellcolor{teal!40}4.52 & 	 \cellcolor{teal!40}4.38	&  4.01&   \cellcolor{teal!40}3.98
\\

\bottomrule[1.2pt]
\end{tabular}
}
    \label{tab: Combining MEIT with a Supervised Approach}
\end{table*}

\bruce{\noindent \textbf{Ablation Study of ECG Encoder} we conducted additional experiments comparing our default 1-D Temporal Convolution ECG encoder with alternative architectures, including: 1. S4-based Model: Vim-B (Vision Mamba, 98M parameters)~\citep{Zhu2024VisionME}. 2. Transformer-based Model: ViT-B/16 (Vision Transformer, 86M parameters)~\citep{Dosovitskiy2020AnII}, adapted for 1-D token patching to align with the temporal nature of ECG signals. 3. SSL-Transformer Model: ViT-B/75 initialized with self-supervised learning (SSL) weights specific to ECG signals~\citep{Na2024GuidingMR}. We evaluated these models on two tasks: Quality of Generated Reports using the MIMIC-IV-ECG dataset, and Evaluation of Alignment with Human Expert Annotations using the PTB-XL dataset. For fair comparison, we used Meta-Llama-3-8B-Instruct as the LLM backbone due to its consistent strong performance.

The results, summarized in the table below, show that our 1-D Temporal Convolution ECG encoder, despite having significantly fewer parameters, performs comparably or better across most metrics compared to ViT and ViT-SSL, and comprehensively outperforms the S4-based Vim. Notably, the ViT-SSL encoder demonstrates the benefit of self-supervised pretraining for initial ECG representation learning. However, our default ECG encoder effectively captures the 12-channel ECG temporal patterns while remaining lightweight, making it well-suited for our efficient instruction tuning framework. These findings validate the effectiveness of our 1-D Temporal Conv encoder and also provide valuable insights for future work, including designing more complex ViT-based architectures optimized for ECG time-series data.}

\begin{table*}[h]
   \centering
\caption{Parameter Comparison of ECG encoder and LLM backbone. We use LLaMA-1 7B as an example.}
\scalebox{0.90}{
\begin{tabular}{lcc}
        \midrule[1.2pt]
        \textsc{Module} &  \textbf{Trainable Params} & \textbf{Inference Params} \\
        \midrule
        LLM backbone &  159M &  6.90B \\
        ECG encoder    & 20.4M & 20.4M    \\
        \midrule[1.2pt]
        \end{tabular}
        \label{tab: params compare}
}
\end{table*}

\subsection{Further Analysis of Generated Prompts}
\label{appendix: Further Analysis of Generated Prompts}

\noindent \textbf{Prompts Number Analysis} In the ECG instruction data curation, we manually created 32 prompt examples, as illustrated in Section~\ref{sec:instruction_data}. To increase the diversity of our samples, we employed GPT-4 to rephrase these manually designed prompts, generating a larger pool of prompt examples. These generated examples were randomly sampled and paired with ECG-text pairs to compile the ECG instruction dataset. In this section, We compare the experiment's effects using 128, 256, and 512-generated samples, respectively. Table~\ref{tab: nums ablaition} shows the corresponding results with different dimensions. When the number is 256, it can achieve
better results in most experimental settings. Hence, we take 256 generated samples as our default setting during the instruction tuning and inference.

\begin{table*}[h]
    \centering
    \caption{Performance comparison of different numbers of generated prompt samples. We evaluate them on the MIMIC-IV-ECG dataset, using BLEU-4, METEOR, ROUGE-L, and CIDEr-D metrics. We take LLaMA-1 7B as the LLM backbone here. 
    \colorbox{teal!40}{heavy teal} color indicates the highest results.}
    \scalebox{0.90}{
    \begin{tabular}{c|cccc}
   \toprule[1.2pt]
\textsc{Prompt Nums}& \textbf{BLEU-4} & \textbf{METEOR} & \textbf{ROUGE-L}  & 
\textbf{CIDEr-D} \\
\midrule[1.2pt]
   128 & 0.541&  0.756& 0.718& 5.15 \\
   256 & \cellcolor{teal!40}0.543& \cellcolor{teal!40}0.761 & 0.724& \cellcolor{teal!40}5.26 \\
   512 & 0.538& 0.754 & \cellcolor{teal!40}0.732& 5.03 \\
\bottomrule[1.2pt]
\end{tabular}
}
    
    \label{tab: nums ablaition}
\end{table*}

\noindent \textbf{Ablation Study on GPT-4 Prompt Rephrasing } \zw{We also conducted an ablation study to compare the performance with and without GPT-4 rephrasing prompts, using a fixed prompt for the latter. The results in the following Table~\ref{tab: ablation GPT-4 Prompt Rephrasing} indicate that using diverse prompts rephrased by GPT-4 leads to better performance, highlighting the superiority of instruction tuning in enhancing LLMs' generalization to new tasks and data over direct fine-tuning. }

\begin{table*}[h]
    \centering
    \caption{Performance comparison of with and without GPT-4 prompt rephrasing. We take Mistral-Instruct as the LLM backbone here. 
    \colorbox{teal!40}{heavy teal} color indicates the highest results.}
    \scalebox{0.90}{
    \begin{tabular}{c|cccc}
   \toprule[1.2pt]
\textsc{Prompt Nums}& \textbf{BLEU-4} & \textbf{METEOR} & \textbf{ROUGE-L}  & 
\textbf{CIDEr-D} \\
\midrule[1.2pt]
   \textit{w.o.}  Rephrasing & 0.564&  0.745& 0.738& 5.50 \\
   \textit{w.}  Rephrasing (Ours) & \cellcolor{teal!40}0.576& \cellcolor{teal!40}0.768 & \cellcolor{teal!40}0.751& \cellcolor{teal!40}5.62 \\
\bottomrule[1.2pt]
\end{tabular}
}
    \label{tab: ablation GPT-4 Prompt Rephrasing}
\end{table*}

\subsection{Comparison with Encoder-Decoder Models}
\label{appendix: Comparison with Encoder-Decoder Models}

\zw{In this section, we conducted additional comparative experiments using two open-source traditional encoder-decoder architectures: BART-Large (406M parameters)~\citep{lewis2019bart} and T5-Large (780M parameters)~\citep{raffel2020exploring}, as shown in Table~\ref{encoder-decoder-based models}. In adapting our framework for ECG instruction tuning, we employ the language encoder to process the input instruction, an ECG encoder to handle the input ECG signals, and the language decoder to generate the ECG report based on the output from both language end ECG encoder.

Our findings indicate that the performance of encoder-decoder models is comparable to the small pre-trained language models (GPT2-Medium and GPT-Large) presented in Table~\ref{tab: res mimic nlg} and Table~\ref{tab: res ptb nlg} of our paper. Moreover, LLM-based backbones (such as LLaMA1-2) consistently achieve a significant margin of improvement over the encoder-decoder architectures across all metrics.}

\begin{table*}[h]
    \centering
    \vspace{-7pt}
    \caption{ Comparison with encoder-decoder-based models on MIMIC-IV-ECG. For model size, 'M' denotes the million level, and 'B' denotes the billion level.
    The \colorbox{teal!15}{light teal} color indicates the second highest results, and
    \colorbox{teal!40}{heavy teal} color indicates the highest results.}
    \label{tab: encoder-decoder-mimic-nlg}
    \scalebox{0.66}{
    \begin{tabular}{c|c|ccccccccc}
   \toprule[1.2pt]
\textsc{Models} & \textsc{Size} & \textbf{BLEU-1} & \textbf{BLEU-2} & \textbf{BLEU-3} & \textbf{BLEU-4} & \textbf{METEOR} & \textbf{ROUGE-L} & \textbf{ROUGE-1} & \textbf{ROUGE-2} & \textbf{CIDEr-D} \\
\midrule[1.2pt]
BART-Large& 406M& 0.525 & 0.498 & 0.466 & 0.388 & 0.455 & 0.472 & 0.5124 & 0.451 & 3.15 \\
T5-Large& 780M & 0.595 & 0.542 & 0.465 & 0.422 & 0.498 & 0.456 & 0.522 & 0.438 & 4.08 \\
\midrule
LLaMA-1 & 7B & \cellcolor{teal!15}0.685 & \cellcolor{teal!15}0.648 & \cellcolor{teal!15}0.615 & \cellcolor{teal!15}0.543 & \cellcolor{teal!15}0.761 & \cellcolor{teal!15}0.724 & \cellcolor{teal!15}0.742 & \cellcolor{teal!15}0.642 & \cellcolor{teal!15}5.26 \\
\midrule
LLaMA-2-Instruct & 7B &\cellcolor{teal!40}0.706 & \cellcolor{teal!40}0.662 & \cellcolor{teal!40}0.622 & \cellcolor{teal!40}0.581 & \cellcolor{teal!40}0.775 & \cellcolor{teal!40}0.745 & \cellcolor{teal!40}0.768 & \cellcolor{teal!40}0.664 & \cellcolor{teal!40}5.55 \\
\midrule[1.2pt]
\end{tabular}
}
\label{encoder-decoder-based models}
\vspace{-15pt}
\end{table*}

\subsection{Analysis of Combining MEIT with a Supervised Manner }
\label{appendix: Analysis of Combining MEIT with a Supervised Approach}

\zw{In this section, we conduct a new experiment where we trained a CNN (ECG encoder) in a supervised manner on the PTB-XL training set, utilizing all available annotations (approximately 70 patterns), as shown in Table~\ref{tab: Combining MEIT with a Supervised Approach}. We then transferred the CNN for ECG instruction fine-tuning on both the MIMIC-IV-ECG and PTB-XL datasets.  Our findings indicate that performance increased on the PTB-XL dataset in most metrics, likely due to the model's prior learning of specific annotated patterns. However, performance fluctuated on the MIMIC-IV-ECG dataset, which contains more data and exhibits greater diversity. This suggests that the supervised approach may enhance performance on in-domain data, but it limits generalizability to data from unseen domains.}

\begin{table*}[h]
    \centering
    \caption{Comparisons of results with and without supervised manner. We take LLaMA-2-Instruct as the LLM backbone here. 
    \colorbox{teal!40}{heavy teal} color indicates the highest results.}
    \scalebox{0.90}{
    \begin{tabular}{c|cccc}
   \toprule[1.2pt]
   \textsc{Methods} & \multicolumn{4}{c}{\textsc{PTB-XL}} \\
   \midrule[1.2pt]
& \textbf{BLEU-4} & \textbf{METEOR} & \textbf{ROUGE-L}  & 
\textbf{CIDEr-D} \\
\midrule[1.2pt]
   MEIT  & 0.439&  \cellcolor{teal!40}0.675& 0.594& 4.05 \\
   MEIT + Supervised manner &  \cellcolor{teal!40}0.445& 0.664 &  \cellcolor{teal!40}0.612 &  \cellcolor{teal!40}4.12 \\
   \midrule[1.2pt]
      & \multicolumn{4}{c}{\textsc{MIMIC-IV-ECG}} \\
   \midrule[1.2pt]
& \textbf{BLEU-4} & \textbf{METEOR} & \textbf{ROUGE-L}  & 
\textbf{CIDEr-D} \\
\midrule[1.2pt]
   MEIT  & \cellcolor{teal!40}0.581 &  0.775 & \cellcolor{teal!40}0.745& \cellcolor{teal!40}5.55 \\
   MEIT + Supervised manner & 0.578 & \cellcolor{teal!40} 0.778 & 0.739 & 5.47 \\
   
\bottomrule[1.2pt]
\end{tabular}
}
    \label{tab: Combining MEIT with a Supervised Approach}
\end{table*}

\subsection{Computational Cost Analysis of MEIT}
\label{appendix: Computational Cost Analysis of MEIT}

\zw{The time cost experiment, detailed in the Table~\ref{tab: cost compare}, was conducted on the MIMIC-IV-ECG dataset. We found that larger models have longer training and inference times. Thus, we are considering techniques like quantization and other compression methods to improve model efficiency in future work.
}

\begin{table*}[h]
   \centering
\caption{Computational time Analysis of MEIT with various parameters and backbones. }
\scalebox{0.90}{
\begin{tabular}{c|c|cc}
        \midrule[1.2pt]
        \textsc{Model} & \textsc{Size} &  \textbf{Training time} & \textbf{Testing  time}  \\
        \midrule
       & & 4 A100 and 3 Epochs & 1 A100 and 128 Generated Samples \\
        \midrule
        GPT-2 Large &774M & 3.25h &  3.125min \\
        LLaMA-2-Instruct  & 7B &  13.5h & 9 min    \\
        LLaMA-2-Instruct (+)  & 13B & 27h &  14.125 min    \\
        \midrule[1.2pt]
        \end{tabular}
        \label{tab: cost compare}
}
\end{table*}

\subsection{Examples of ECG Instruction Tuning Data and the Corresponding Generated ECG Reports}
\label{appendix: sample visualization}

\begin{figure*}[h]
\centering
\includegraphics[width=0.66\textwidth]{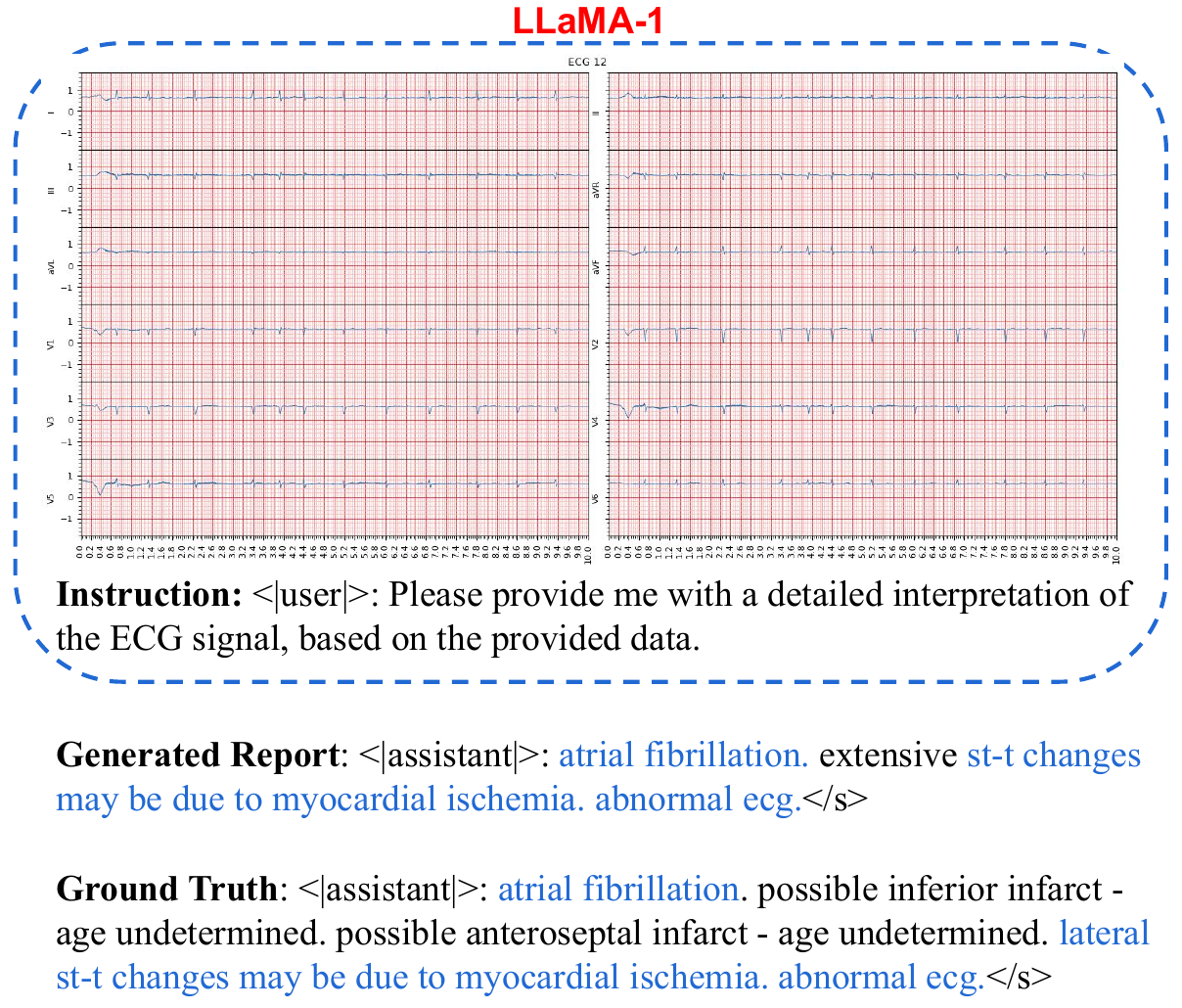}
\caption{Reports generated by LLaMA-1 following ECG instruction Tuning.}
\label{fig:llama1_ecg}
\end{figure*}

\begin{figure*}[h]
\centering
\includegraphics[width=0.66\textwidth]{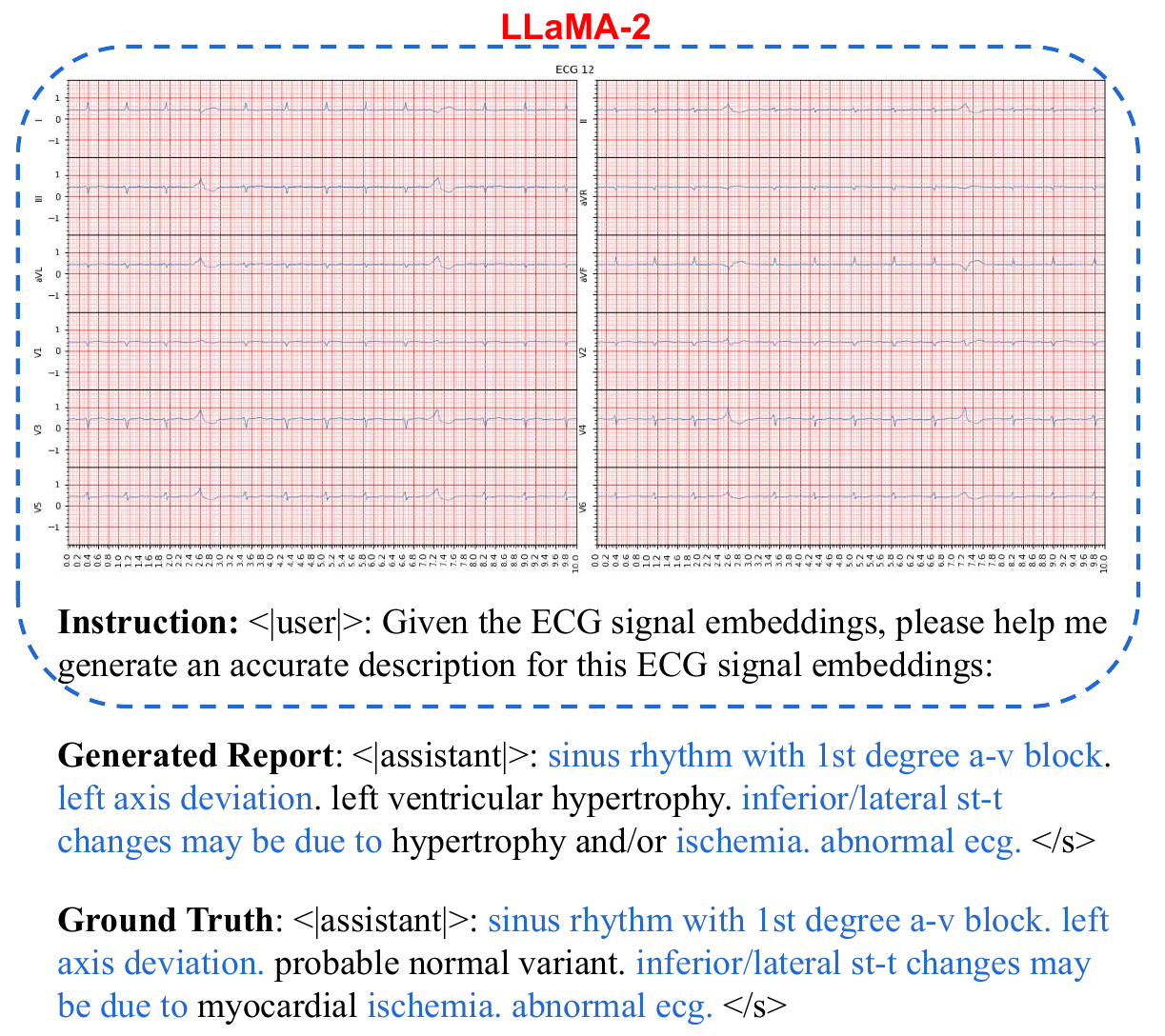}
\caption{Reports generated by LLaMA-2 following ECG instruction Tuning.}
\label{fig:llama2_ecg}
\end{figure*}

\begin{figure*}[h]
\centering
\includegraphics[width=0.66\textwidth]{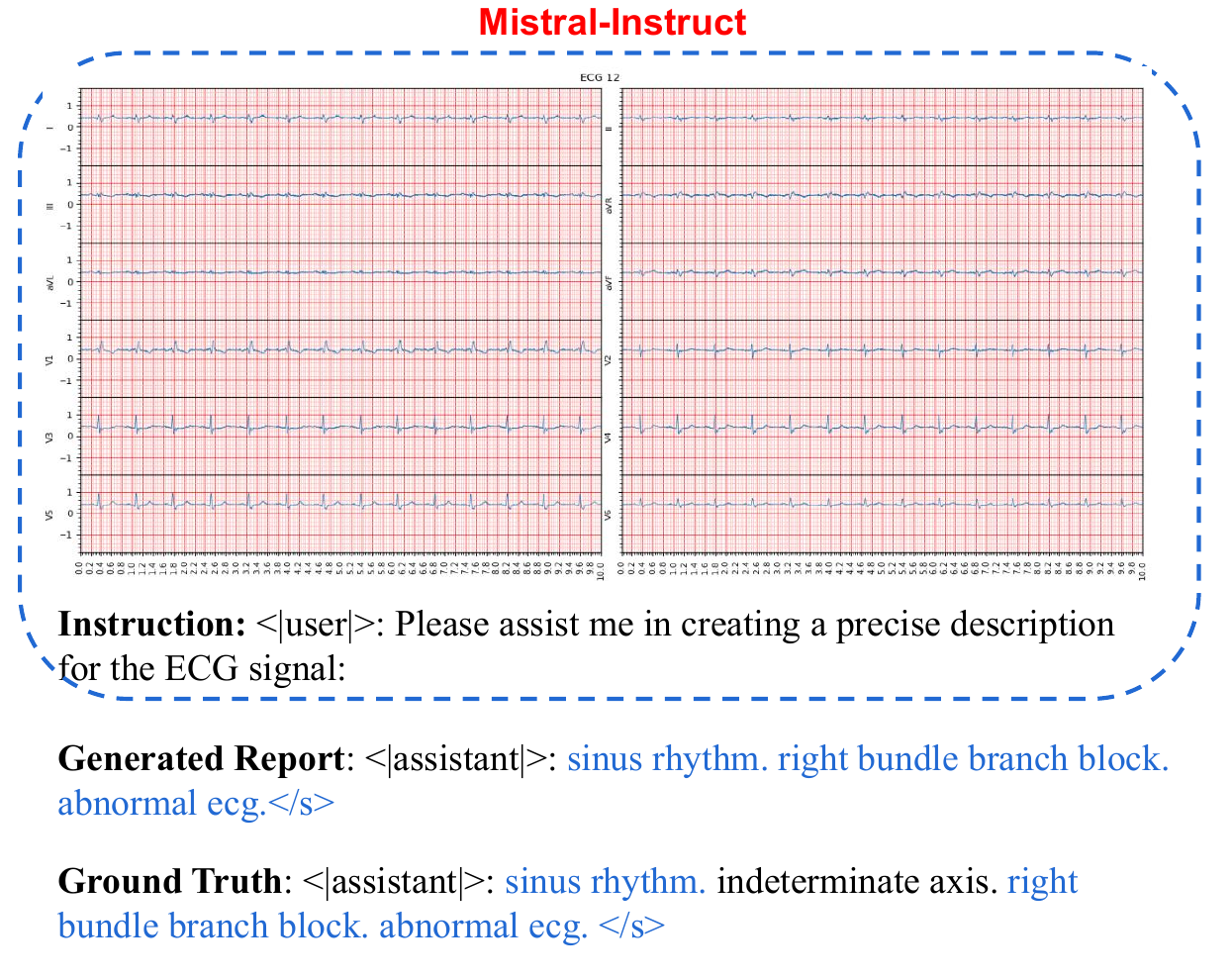}
\caption{Reports generated by Mistral-Instruct following ECG instruction Tuning.}
\label{fig:mistral_instruct}
\end{figure*}

As illustrated in Figures~\ref{fig:llama1_ecg}, ~\ref{fig:llama2_ecg}, and ~\ref{fig:mistral_instruct} we have visualized the report samples generated by LLaMA-1, LLaMA-2, and Mistral-Instruct. The samples are presented in \blue{blue} font to highlight the key information that aligns with the ground truth. The visualization demonstrates that all three selected models can capture the essential patterns of ECG signals and generate accurate reports. This underscores the efficacy of our proposed \textbf{MEIT} framework, which is adaptable to most open-source LLMs. It effectively learns the correct clinical semantics of ECG signals, thereby enabling the generation of corresponding reports.

\subsection{Future Works}

In this work, we propose an effective and efficient method for generating ECG reports. Since this is a generative task, the content produced may occasionally lead to hallucinations~\cite{zhu2025}. This is particularly pertinent in medical contexts such as ECG report generation, where accuracy is of utmost importance. To mitigate this issue, future iterations of our work could incorporate retrieval-augmented methods~\cite{wan2022g, li2024uncertaintyrag}, information optimized strategies~\cite{xiong2022self, liang2020large, liang2020many}, multi-modal fusion and
optimization strategies~\cite{wan2024look, xiong2024autoregressive, liu2024benchmarking, gong2024neuroclips, wan2023spatio, chen2025recent, luo2024enhancing, shen2025efficient, wan2025meda, zhang2025enhancing, liu2024can, huang2024evolver, zhu2024dglf, liu2025knowledge}, Reasoning enhanced methods~\cite{wan2025srpo, liu2025beyond, shen2025phyx}, efficient ML~\cite{wang2024svd, tao2024scaling, wan2024d2o, liu2024contemporary, xin2024v, shen2024famba, wang2025svd, xiong2024uncomp, xiong2025parallelcomp}. For instance, by integrating a retrieval mechanism that accesses a database of verified medical or biological knowledge~\cite{zheng2024structure} or previous ECG reports, the model could be guided towards generating more accurate and reliable outputs.

\end{document}